\documentclass{article}

\usepackage{microtype}
\usepackage{graphicx}
\usepackage{subcaption}
\usepackage{booktabs} % for professional tables

\usepackage{hyperref}

\usepackage[preprint]{icml2026}

\usepackage{amsmath}
\usepackage{amssymb}
\usepackage{mathtools}
\usepackage{amsthm}

\usepackage[utf8]{inputenc} % allow utf-8 input
\usepackage[T1]{fontenc}    % use 8-bit T1 fonts
\usepackage{url}            % simple URL typesetting
\usepackage{amsfonts}       % blackboard math symbols
\usepackage{nicefrac}       % compact symbols for 1/2, etc.
\usepackage{microtype}      % microtypography
\usepackage{xcolor}         % colors
\usepackage{color}
\usepackage{multirow}
\usepackage[table]{xcolor} 
\usepackage[normalem]{ulem}
\usepackage{wrapfig}

\usepackage[capitalize,noabbrev]{cleveref}

\theoremstyle{plain}

\theoremstyle{definition}

\theoremstyle{remark}

\usepackage[textsize=tiny]{todonotes}

\icmltitlerunning{LightMoE: Reducing Mixture-of-Experts Redundancy through Expert Replacing}

\begin{document}

\twocolumn[
  % \icmltitle{Submission and Formatting Instructions for \\
  %   International Conference on Machine Learning (ICML 2026)}
  % \icmltitle{LightMoE: Reducing Mixture-of-Experts Redundancy \\ through Hierarchical Parameter Sharing}
\icmltitle{LightMoE: Reducing Mixture-of-Experts Redundancy\texorpdfstring{\\}{ }through Expert Replacing}

  % It is OKAY to include author information, even for blind submissions: the
  % style file will automatically remove it for you unless you've provided
  % the [accepted] option to the icml2026 package.

  % List of affiliations: The first argument should be a (short) identifier you
  % will use later to specify author affiliations Academic affiliations
  % should list Department, University, City, Region, Country Industry
  % affiliations should list Company, City, Region, Country

  % You can specify symbols, otherwise they are numbered in order. Ideally, you
  % should not use this facility. Affiliations will be numbered in order of
  % appearance and this is the preferred way.
  \icmlsetsymbol{equal}{*}

  \begin{icmlauthorlist}
    \icmlauthor{Jiawei Hao}{bit}
    \icmlauthor{Zhiwei Hao}{chk}
    \icmlauthor{Jianyuan Guo}{chk}
    \icmlauthor{Li Shen}{zhongshan}
    \icmlauthor{Yong Luo}{wuhan}
    \icmlauthor{Han Hu}{bit}
    \icmlauthor{Dan Zeng}{shanghai}
    %\icmlauthor{}{sch}
    %\icmlauthor{}{sch}
    %\icmlauthor{}{sch}
  \end{icmlauthorlist}

  \icmlaffiliation{bit}{School of Information and Electronics, Beijing Institute of Technology, Beijing, China}
  \icmlaffiliation{chk}{Department of Computer Science, City University of Hong Kong, Hong Kong, China}
  \icmlaffiliation{wuhan}{School of Computer Science, Wuhan University, Wuhan, China}
  \icmlaffiliation{zhongshan}{Shenzhen Campus of Sun Yat-sen University, Shenzhen, China}
  \icmlaffiliation{shanghai}{School of Communication and Information Engineering, Shanghai University, Shanghai, China}

  \icmlcorrespondingauthor{Han Hu}{hhu@bit.edu.cn}

  % You may provide any keywords that you find helpful for describing your
  % paper; these are used to populate the "keywords" metadata in the PDF but
  % will not be shown in the document
  \icmlkeywords{Machine Learning, ICML}

  \vskip 0.3in
]

% this must go after the closing bracket ] following \twocolumn[ ...

% This command actually creates the footnote in the first column listing the
% affiliations and the copyright notice. The command takes one argument, which
% is text to display at the start of the footnote. The \icmlEqualContribution
% command is standard text for equal contribution. Remove it (just {}) if you
% do not need this facility.

% Use ONE of the following lines. DO NOT remove the command.
% If you have no special notice, KEEP empty braces:
\printAffiliationsAndNotice{}  % no special notice (required even if empty)
% Or, if applicable, use the standard equal contribution text:
% \printAffiliationsAndNotice{\icmlEqualContribution}

\newcommand{\MethodName}{LightMoE}

\begin{abstract}
Mixture-of-Experts (MoE) based Large Language Models (LLMs) have demonstrated impressive performance and computational efficiency. However, their deployment is often constrained by substantial memory demands, primarily due to the need to load numerous expert modules. While existing expert compression techniques like pruning or merging attempt to mitigate this, they often suffer from irreversible knowledge loss or high training overhead.
In this paper, we propose a novel expert compression paradigm termed expert replacing, which replaces redundant experts with parameter-efficient modules and recovers their capabilities with low training costs. We find that even a straightforward baseline of this paradigm yields promising performance. Building on this foundation, we introduce \MethodName{}, a framework that enhances the paradigm by introducing adaptive expert selection, hierarchical expert construction, and an annealed recovery strategy.
Experimental results show that \MethodName{} matches the performance of LoRA fine-tuning at a 30\% compression ratio. Even under a more aggressive 50\% compression rate, it outperforms existing methods and achieves average performance improvements of 5.6\% across five diverse tasks. These findings demonstrate that \MethodName{} strikes a superior balance among memory efficiency, training efficiency, and model performance.

\end{abstract}

\section{Introduction}

LLMs leveraging Sparse Mixture-of-Experts (MoE) architectures, exemplified by models such as DeepSeek-MoE \cite{deepseekmoe, deepseekv2} and OLMoE \cite{OLMoE}, have recently received significant attention. Their design offers excellent performance and notable efficiency in both training and inference processes.
However, a primary challenge associated with these powerful models is their substantial memory footprint. Loading numerous expert modules demands considerable memory resources, which constrains their practical applicability and impedes widespread deployment in real-world scenarios.
While expert offloading techniques \cite{Kim2024ES,Eliseev2023FastInference,Yu2025fMoE} mitigate GPU memory limits, they introduce prohibitive inference latency due to the frequent transfer of weights from CPU memory or disk. Consequently, direct parameter compression has become a critical research frontier.

Current MoE compression methods largely follow two paradigms: expert pruning \cite{Lu2024NAEE, yang2024moe_i^2} and expert merging \cite{mcsmoe, Liu2024EEP, hcsmoe}. Expert pruning methods, such as MoE-Pruner, which prunes weights based on activation frequencies and router importance scores, aim to remove less critical experts. However, a significant drawback of pruning is the irreversible loss of pruned expert knowledge, which results in substantial performance degradation. On the other hand, expert merging seeks to combine multiple experts into a single, more compact representation. However, merging experts inherently diminishes the model’s representational diversity, and determining an optimal merging strategy remains a significant challenge in the field. Despite these challenges, recent findings \cite{LiuD0PCCT24,Lu2024NAEE} reveal that current MoEs contain significant parameter redundancy. Some studies \cite{esft} also indicate a high concentration of active experts in fine-grained MoE models when applied to specific tasks.

% \begin{figure}[ht]
%   % \vskip 0.2in
%   \begin{center}
%     \centerline{\includegraphics[width=0.48\textwidth]{src/comparison_new.pdf}}
%     \caption{Removal of less important experts lead to significant performance decrease.}
%     \label{fig:comparison}
%   \end{center}
% \end{figure}

\begin{figure}[!ht]
  \centering
\includegraphics[width=0.45\textwidth]{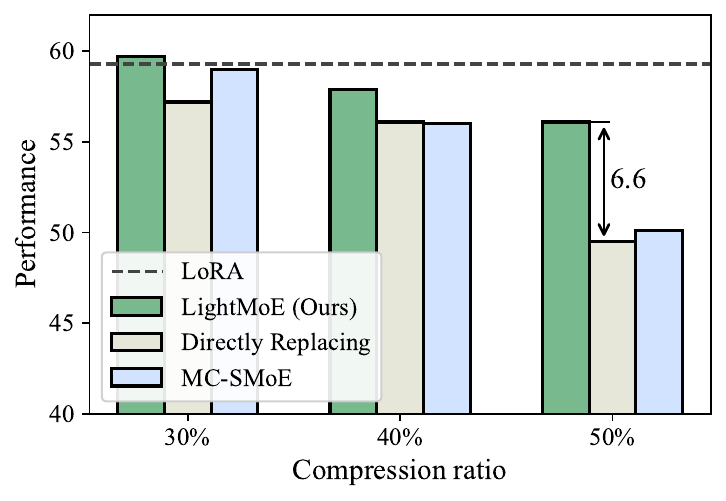}
  \caption{Performance comparison across different compression ratios on the Math task. While the directly replacing strategy performs comparably to MC-SMoE, both suffer from significant performance degradation, particularly at higher compression ratios}
  \label{fig:comparison}
  % \vspace{-0.6cm}
\end{figure}

Based on these challenges and observations, we explore a straightforward solution: \textit{Can we simply replace less critical experts with parameter-efficient modules and subsequently recover their capabilities with low-cost training}? 
This idea raises three fundamental questions:
(1) \textit{expert selection}: How to select the less important experts for replacement?
(2) \textit{module construction}: How to design and initialize the parameter-efficient modules?
(3) \textit{efficient recovery}: How to restore model performance with minimal training overhead? A simple way to materialize this idea is to replace these less frequently activated experts with Low-Rank Adaptation (LoRA) \cite{hu2022lora}, and subsequently fine-tune the modified model. As illustrated in \cref{fig:comparison}, this directly replacing strategy achieves performance comparable to the existing method MC-SMoE \cite{mcsmoe}. However, both approaches suffer from significant performance degradation, particularly at higher compression ratios. This suggests that even experts deemed ``inactive'' or less critical for a specific task may still harbor fundamental abilities and knowledge crucial for overall model capabilities. Attempting to restore this lost knowledge and alleviate performance degradation solely through fine-tuning proves to be a difficult endeavor.
% Restoring this lost knowledge and alleviating performance degradation solely through fine-tuning proves to be a difficult endeavor.

In this work, we propose \MethodName, a framework for compressing redundant experts in MoE models via expert replacing. The framework comprises three key stages: selecting less important experts through adaptive thresholding, constructing hierarchical experts, and progressively replacing the original experts through annealing. First, LightMoE calculates the relative importance of experts within and across layers to establish an adaptive threshold. This threshold serves as a dynamic criterion to select less critical experts. Next, the selected experts are grouped and replaced with a smaller number of shared bases, each equipped with task-specific low-rank adaptation parameters to preserve specialization. Finally, during the fine-tuning phase, the original experts are gradually annealed into their corresponding shared bases, allowing the model to adapt smoothly to the compressed structure.
Experimental results across five diverse tasks show that \MethodName{} achieves performance comparable to LoRA fine-tuning at a 30\% compression ratio. Even under a more aggressive 50\% compression rate with the same training budget, it delivers average performance improvements of 5.6\% and 3.8\% over existing methods and the directly replacing baseline, respectively.
% Even under a more aggressive 50\% compression rate, it delivers an average performance improvement of 5.6\% over existing methods under the same training budget (not compare replacing).

\section{Related Works}

% \textbf{Mixture-of-Experts LLMs.}
\subsection{Mixture-of-Experts LLMs}
In contrast to traditional dense Large Language Models (LLMs) such as the LLaMA series \cite{touvron2023llama, grattafiori2024llama}, Mixture of Experts (MoE) architectures offer a compelling approach to expanding model capacity while reducing training and inference costs. The majority of existing MoE \cite{lepikhin2020gshard, dai2022stablemoe, shen2024jetmoe} implementations employ coarse-grained architectures with relatively few experts. For instance, the Mixtral series \cite{jiang2024mixtral} activate only 2 out of 8 available experts during computation. This limitation necessitates that each expert must handle diverse patterns across multiple domains simultaneously. 

More recently, fine-grained expert segmentation \cite{deepseekmoe,qwen2,qwen2.5} has gained significant attention, as it enables a greater variety of expert combinations and demonstrates superior performance. In the OLMoE \cite{OLMoE}, there are as many as 64 experts, with 8 active experts. However, these fine-grained MoE architectures inherit substantial memory footprints from storing all expert weights.
Interestingly, recent studies \cite{esft} reveal that this fine-grained division ensures a high degree of specialization among the experts. We argue that this distinct specialization presents opportunities for task-specific model compression, enabling more efficient deployment for downstream applications.
% More recently, fine-grained expert segmentation \cite{deepseekmoe,qwen2,qwen2.5} has gained significant attention, as it enables a greater variety of expert combinations and demonstrates superior performance. In the OLMoE \cite{OLMoE}, there are as many as 64 experts, with 8 active experts. Recent studies \cite{esft} shows that the fine-grained division of experts ensures a high degree of specialization among the experts. However, these fine-grained MoE architectures inherit substantial memory footprints from storing all expert weights. Interestingly, the high degree of expert specialization presents promising opportunities for task-specific model compression, enabling more efficient deployment for downstream applications.

\subsection{MoE Compression Methods}

Given that experts constitute a significant portion of memory requirements, prior work on MoE model efficiency can be categorized into these principal approaches: expert offloading \citep{Kim2024ES,Eliseev2023FastInference,Yu2025fMoE}, expert compression \cite{Lu2024NAEE, yang2024moe_i^2, mcsmoe, Liu2024EEP}, and more general compression strategies such as quantization \cite{Li2024Quant,Huang2024MCMoE} and knowledge distillation \cite{xu2024distillation, Kim2025Every}. 

In this work, we focus on the expert compression technique, which can be broadly categorized into expert pruning \cite{Lu2024NAEE, yang2024moe_i^2} and expert merging \cite{mcsmoe, Liu2024EEP, hcsmoe}.
Expert pruning methods focus on pruning dense matrices and removing redundant experts. For instance, MoE-Pruner \cite{xie2024moepruner} prunes weights based on activation frequencies, while NAEE \cite{Lu2024NAEE} searches for expert combinations that minimize output deviation. However, these approaches often lead to substantial performance degradation due to the permanent loss of expert knowledge. Furthermore, search-based methods like NAEE are ill-suited for fine-grained MoE architectures due to combinatorial explosion. In the case of OLMoE with 64 experts, a 50\% reduction results in an overwhelming search space of $C(64,32)\approx 1.8\times 10^{18}$ combinations per layer.
Expert merging methods aim to consolidate multiple experts into fewer ones. For example, MC-SMoE \cite{mcsmoe} clusters experts according to their routing statistics and subsequently merges each cluster into a single representative expert. However, this approach inherently diminishes the model's representational diversity, and identifying an optimal merging strategy is non-trivial. Furthermore, while MC-SMoE employs progressive low-rank decomposition during retraining for further expert compression, it introduces substantial training overhead. This is because the decomposition process requires computing gradients from the original experts.
Recent efforts have also explored dynamic inference optimization. Specifically, \citet{he2025efficiently_edit} propose replacing auxiliary activated experts with compressed modules to reduce active parameters. While this effectively lowers inference computational costs, it does not address the substantial memory footprint stemming from the total parameters, as the non-active experts still require storage.

Distinct from the aforementioned approaches, we propose a novel expert compression paradigm termed \textit{expert replacing}. Empirical results show that even a simple baseline of this paradigm achieves performance comparable to, or slightly superior to, existing methods.
Building on this foundation, we further propose optimizations across three key dimensions: expert selection, module construction, and efficient recovery. This design not only enables effective model compression but also preserves the specialized capability of the model with minimal overhead. Consequently, it strikes an optimal balance among memory efficiency, training efficiency, and model performance.
From a broader perspective, expert merging can be viewed as a special case of our expert replacing paradigm. Fundamentally, merging does not reduce the number of experts indexed by the router, but instead maps multiple indices to the same parameters. In essence, this is equivalent to replacing multiple original experts with a single shared expert.
% In this work, we focus on the expert compression technique, which can be broadly categorized into expert pruning \cite{Lu2024NAEE, yang2024moe_i^2} and expert merging \cite{mcsmoe, Liu2024EEP, hcsmoe} methods. Expert pruning methods focus on pruning dense matrices. For example, MoE-Pruner \cite{xie2024moepruner} prunes weights of experts based on their activations and router importance to achieve compression. However, these approaches they often result in substantial performance degradation due to the loss of expert knowledge. 
% Expert merging methods aim to consolidate multiple experts into fewer experts. For example, MC-SMoE \cite{mcsmoe} clusters experts according to their routing statistics and subsequently merges each cluster into a single representative expert. However, merging experts inherently diminishes the model's representational diversity and determining an optimal merging strategy remains a significant challenge in the field.(descript MC-SMoE compress method, and show it require large training cost.)
% Distinct from the aforementioned approaches, we propose a novel expert compression paradigm termed expert replacing. We replace multiple less important experts with a set of shared experts and equip each expert with low-rank adaptation parameters. This design not only enables effective model compression but also preserves the specialized capability of the model, striking a balance between memory efficiency and performance.

\subsection{Low-Rank Adaptation}
The immense complexity and computational demands of LLMs with billions of parameters pose substantial challenges for their adaptation to specific downstream tasks. Parameter-efficient fine-tuning (PEFT) \cite{xu2023peft_survey, han2024peft_survey} aims to minimize the fine-tuning parameters while achieving performance comparable to full fine-tuning. Typical PEFT methods include adapter \cite{adapterdrop, condiadapter, wang2022adamix}, soft prompt \cite{li2021prefix,lester2021power, wang2023multitask}, and low-rank adaptation \cite{hu2022lora, pissa, dora, wu2024mixture, lora+}. LoRA \cite{hu2022lora} decomposes the original weight matrices into low-rank components. Inspired by this, we propose replacing multiple experts with parameter-efficient modules, which effectively reduce the number of expert parameters while preserving the diversity of the experts. Moreover, our approach incurs significantly lower training overhead compared to full fine-tuning the entire compressed model.

\section{Method}

% \begin{figure*}
%     \centering
%     \includegraphics[width=\textwidth]{src/main.pdf}
%     \caption{Overview of the proposed \MethodName{} framework. (\textit{Left}) A standard MoE layer. (\textit{Right}) The \MethodName{} workflow, comprising four key steps: (1) expert scoring, (2) selection of low-scoring experts as compression candidates, (3) candidate grouping, and (4) replacement of each group with a shared expert augmented by lightweight expert-specific adaptation parameters.}
%     \label{fig:main}
%     % \vspace{-0.4cm}
% \end{figure*}

\begin{figure*}[ht]
    \centering
    \includegraphics[width=0.95\textwidth]{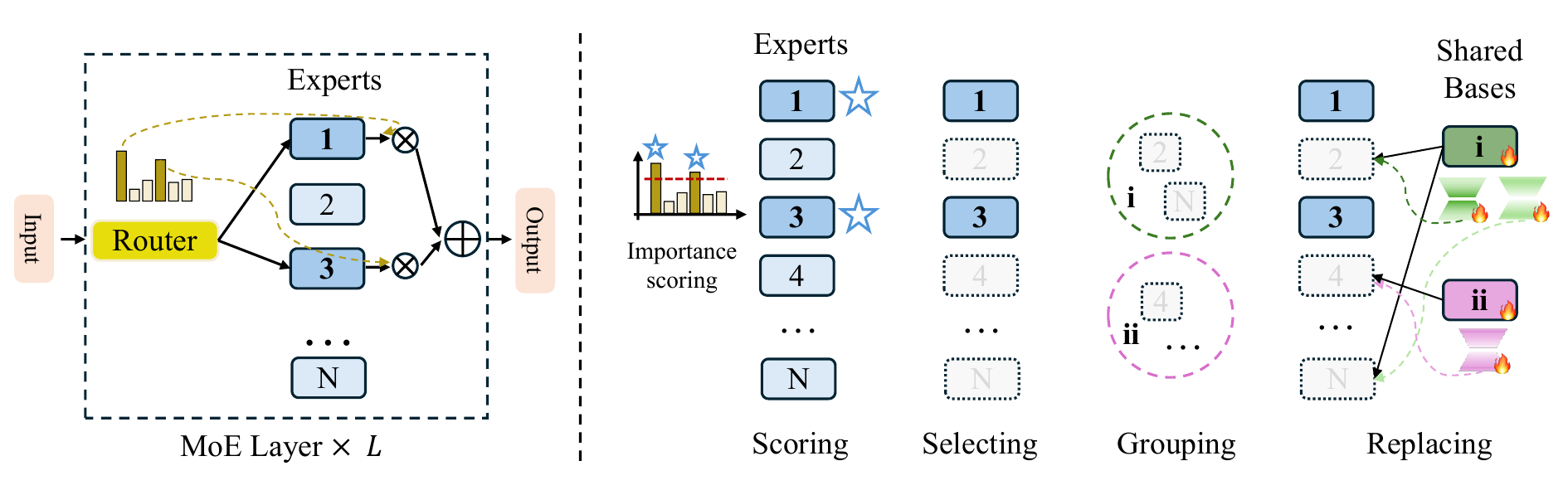}
    \caption{Overview of the proposed \MethodName{} framework. (\textit{Left}) A standard MoE layer. (\textit{Right}) The \MethodName{} workflow, comprising three key steps: (1) scoring experts and selecting those with lower scores as compression candidates, (2) grouping the selected candidates, and (3) replacing each group with a shared base augmented with lightweight, expert-specific adaptation parameters.}
    \label{fig:main}
    % \vspace{-0.4cm}
\end{figure*}

In a MoE layer, a router module selects one or more experts to process each input based on its routing decision, and the outputs of the selected experts are aggregated to form the output. To enable redundant expert compression, the proposed \MethodName{} framework first measures the importance of each expert and selects less critical candidates for compression (\cref{sec:scoring}). These candidate experts are then grouped, and each group is assigned a shared base equipped with low-rank adaptation parameters to retain the specialization of the original experts (\cref{sec:grouping}). Finally, during training, the original candidates are gradually replaced with their corresponding shared bases through an annealed replacement strategy (\cref{sec:replacing}). \cref{fig:main} provides an overview of the proposed framework.

\subsection{Adaptive Expert Selection}
\label{sec:scoring}

In MoE models, redundancy manifests in two dimensions: intra-layer imbalance, where some experts contribute significantly less than others, and inter-layer variability, where different layers exhibit varying degrees of impact on model performance. Building on these observations, we define expert importance based on activation frequency and propose an adaptive strategy to select redundant experts. Specifically, this process involves two steps: computing the importance score for each expert, and identifying less important experts in each layer using an adaptive threshold that incorporates both expert-level and layer-level importance.
% In an MoE model, individual experts do not contribute equally, where some are activated more frequently than others. Building on this observation, we define expert importance based on activation frequency, assuming that more frequently activated experts are more important. Once the importance scores are determined, we select the less important experts for expert compression. This select process consists of two main steps: computing the importance scores of experts and selecting target experts for compression using adaptive thresholding.

\textbf{Importance scoring.}
% In a typical MoE module, multiple feed-forward network (FFN) experts are employed, and a routing mechanism determines which experts are activated for each input token. The output of an MoE layer is computed as:
A typical MoE layer comprises multiple feed-forward network experts and a routing mechanism that dynamically selects the active experts for each input token. The output of the MoE layer is computed as:
\begin{equation}
    y = \sum_{i=1}^{N} G(x)_i E_i(x),
\end{equation}
where $x$ and $y$ denote the input and output of the MoE layer, respectively, $N$ is the total number of experts, $E_i$ represents the $i$-th expert, and $G(x)_i$ is the gating value assigned to expert $i$, indicating its relevance to the input token $x$.

To quantify the importance of each expert, we sample a subset of the training data and aggregate gate values across all tokens. This provides a data-driven estimate of each expert's contribution. To normalize these contributions and enable comparison across experts, we define the normalized gate score for expert $i$ as:
\begin{equation}
    G_i = \frac{\sum_{x\in\mathcal{X}}G(x)_i}{\sum_{i=1}^{N}\sum_{x\in\mathcal{X}}G(x)_i},
\end{equation}
where $\mathcal{X}$ denotes the evaluation subset. This score reflects the relative importance of each expert based on its cumulative gate activation across the sampled data.

\begin{figure}[!t]
\centering
\subfloat[Sorted expert-wise normalized gate scores. Experts exhibit varying importance across layers, highlighting \textit{intra-layer} differences in specialization.]{
    \label{fig:distribution}
    \includegraphics[width=0.37\textwidth]{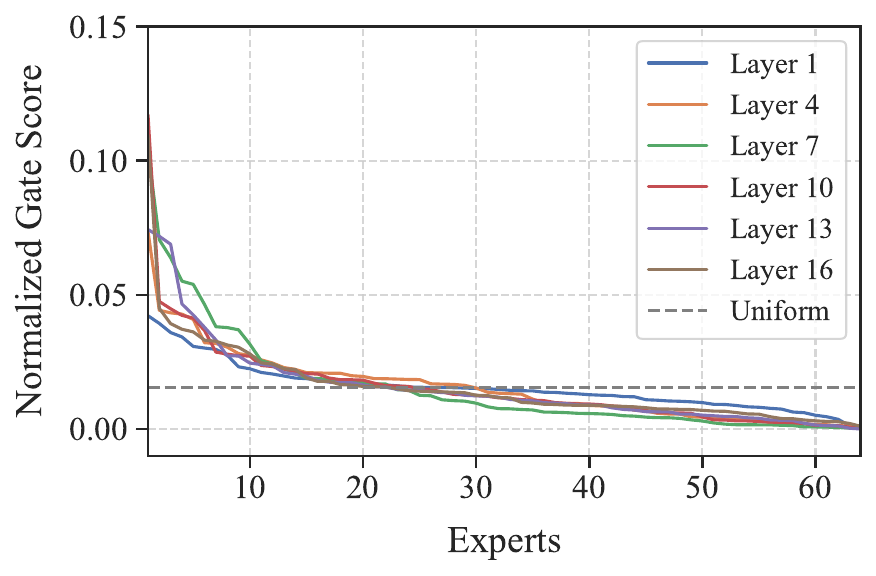}
}
\hfill
\subfloat[Average output norms of router per layer. The increasing norm with depth suggests growing \textit{inter-layer} importance in deeper layers.]{
    \label{fig:norm}
    \includegraphics[width=0.37\textwidth]{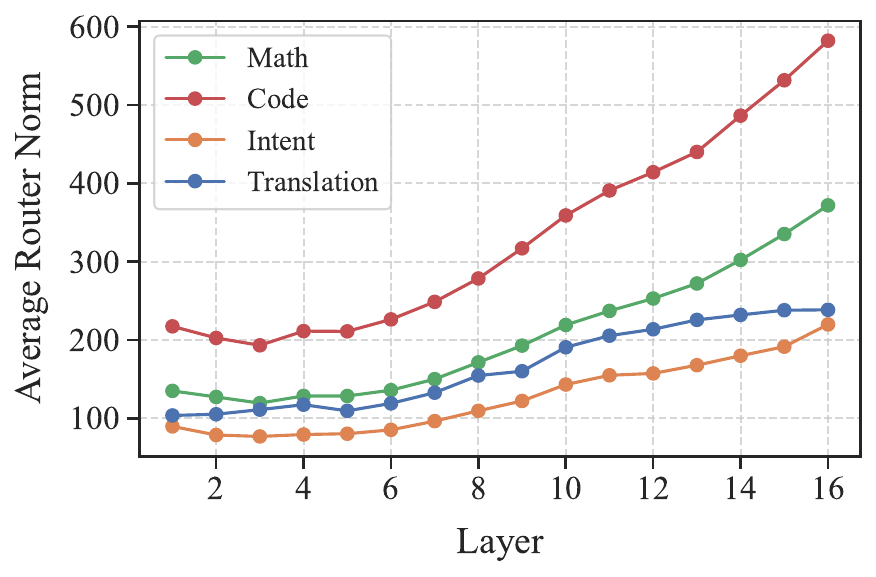}
}
\caption{Analysis of expert importance in OLMoE-1B-7B-SFT.}
\label{fig:trade-off}
% \vspace{-0.4cm}/
\end{figure}

In our preliminary experiment, we used samples from the training set, totaling $2^{17}$ tokens, to evaluate expert importance. The MoE model used is OLMoE-1B-7B-SFT \cite{OLMoE}. \cref{fig:distribution} presents the sorted importance scores of experts in the selected layers. The non-uniform distribution of scores confirms the existence of differences in expert importance, supporting the compression of less important experts in our proposed method.

\textbf{Adaptive thresholding.}
The importance scoring results reveal that experts within the same layer exhibit varying degrees of significance. This suggests that less important experts can be compressed without substantial performance degradation. A direct approach might apply a fixed compression ratio across all experts. However, such a coarse strategy risks unnecessary performance loss.

Motivated by the observed variation in importance among experts within a layer, we hypothesize that similar differences exist across layers in MoE models. To test this, we evaluate the average output norms of router at each layer using a subset of training samples. The underlying intuition is that more important layers tend to produce stronger activations, which reflected in larger output norms. Similar ideas have also been adopted in prior studies~\cite{song2024layer}.
Figure~\ref{fig:norm} presents the results, which show that as the model depth increases, the average router norm also rises, suggesting that deeper layers play a more critical role. Building on this insight, we propose a compression strategy that varies across layers, where shallower layers are assigned higher compression ratios, while deeper layers are preserved more conservatively.
Specifically, based on the previously defined normalized gating scores, we sort experts in ascending order and select the top subset whose cumulative score just exceeds a layer-specific compression threshold. This subset is treated as less important and subject to compression.
Formally, given a base threshold $\hat{p}$, we define the threshold for the $j$-th layer as:
\begin{equation}
\label{equ:adaptive threshold}
    \hat{p}_j=\mathrm{clip}\left( \hat{p} \cdot \mathrm{e}^{-\alpha \left( \mathrm{norm}_j-1 \right)}, p_{\min},p_{\max} \right),
\end{equation}
where $\mathrm{norm}_j$ is the ratio of the norm of layer $j$ to the average norm across all layers. Thus, if the norm of layer $j$  exceeds the mean, the exponential term drives $\hat p_j$ below the base threshold $\hat{p}$. $\alpha$ is an exponential decay coefficient that modulates the strength of the adjustment; we set $\alpha=0.3$ in all experiments. $\mathrm{clip}(p,p_{\min},p_{\max})$ truncates the adjusted threshold to lie within $[p_{\min},\,p_{\max}]$ for numerical stability. In our experiments we set $p_{\min}=0.8\hat{p}$ and $p_{\max}=1.2\hat{p}$.
This adaptive thresholding mechanism allows the proposed method to account for both intra-layer and inter-layer variations in expert importance, resulting in a more effective compression approach.

\subsection{Hierarchical Expert Construction}
\label{sec:grouping}

After selecting less important experts in each layer, we aim to compress them to reduce memory consumption. However, directly removing or merging them may lead to a loss of expert-specific knowledge. To solve this, we propose a hierarchical expert representation that decomposes each expert into a shared base and a expert-specific low-rank adapter. This design allows the model to capture common patterns via the shared base while preserving specific capabilities through efficient parameterization.
Specifically, let $N'$ denote the number of experts to be compressed in a given layer, where each expert is parameterized by a weight matrix $W_1, W_2, \dots, W_{N'} \in \mathbb{R}^{n \times m}$. We first construct the shared base $W_\text{share}$ as the weighted average of these experts, utilizing their normalized gate scores $G_i$ as weights\footnote{For simplicity, we represent each expert using a single weight matrix. In practice, all parameter matrices in each expert are processed in the same manner.}:
\begin{equation}
W_\text{share} = \frac{\sum_{i=1}^{N'} G_i W_i}{\sum_{i=1}^{N'} G_i}.
\end{equation}

Subsequently, each original expert $W_{n'}$ is reconstructed hierarchically as $W_\text{share} + B_{n'} A_{n'}$, where $B_{n'} \in \mathbb{R}^{n \times r}$ and $A_{n'} \in \mathbb{R}^{r \times m}$ represent the expert-specific low-rank adaptation terms, with $r \ll \min(n, m)$.
% After selecting less important experts in each layer, we aim to compress them to reduce memory consumption. However, directly removing or merging them may lead to a loss of expert-specific knowledge. To preserve this knowledge while still achieving compression, we propose a shared expert framework augmented with low-rank adaptation parameters. This design captures essential patterns from less important experts while enabling efficient parameterization.
% Specifically, let $N'$ denote the number of experts to be compressed in a given layer, with each expert parameterized by a weight matrix $W_1, W_2, \dots, W_{N'} \in \mathbb{R}^{n \times m}$. We construct a shared expert weight matrix as a weighted average of these expert parameters, using their normalized gate scores $G_i$ as weights\footnote{For simplicity, we represent each expert using a single weight matrix. In practice, all parameter matrices in each expert are processed in the same manner.}:

% \begin{equation}
% W_\text{share} = \frac{\sum_{i=1}^{N'} G_i W_i}{\sum_{i=1}^{N'} G_i}.
% \end{equation}

% Each compressed expert $W_{n'}$ is then replaced with the combination $W_\text{share} + B_{n'} A_{n'}$, where $B_{n'} \in \mathbb{R}^{n \times r}$ and $A_{n'} \in \mathbb{R}^{r \times m}$, with $r \ll \min(n, m)$. This low-rank adaptation mechanism is inspired by LoRA~\cite{hu2022lora}, and both $B_{n'}$ and $A_{n'}$ are initialized following the same rule.

While a single shared base can serve as a common foundation, it may not adequately capture the diverse knowledge embedded across multiple less important experts. To address this, we extend the framework to support multiple shared bases. Suppose we choose to use $M$ shared bases, where $M < N'$. The $N'$ experts are partitioned into $M$ groups, each represented by a shared base. Following the approach in \cite{mcsmoe}, we select the top $M$ experts with the highest normalized gate scores $G_i$ as dominant experts. Each remaining expert is then assigned to the dominant expert with which it shares the highest semantic similarity. This similarity is computed as the average sample-wise cosine similarity between the routing logits of the non-dominant and dominant experts, using the same evaluation samples from the scoring phase. Once grouped, each set of experts is compressed into a shared base following the weighted average procedure described above.

\textbf{Expert-Level Compression Ratio.}
Under the proposed scheme, multiple full-rank experts are represented using a smaller number of full-rank shared bases, each augmented with several low-rank adaptation terms.
The resulting expert compression ratio $\rho$ is given by:
\begin{equation}
    \rho = 1- \frac{(N - N' + M)nm + N'r(n+m)}{N \cdot nm}.
\end{equation}
Since $M < N'$ and $r \ll \min(n, m)$, this strategy leads to a substantial reduction in parameter count while preserving model capacity and diversity.

\subsection{Annealed Expert Replacement}
\label{sec:replacing}

Directly replacing the original experts with shared bases and low-rank adaptation parameters can lead to significant performance degradation due to the abrupt change in the model parameter space. To mitigate this, we introduce an annealed expert replacement strategy, which gradually transitions each expert from its original form to its compressed representation during fine-tuning. This smooth transition helps preserve performance by maintaining continuity in the optimization trajectory. A detailed empirical analysis of the training dynamics is provided in appendices (\cref{sec:training loss}).

Concretely, consider a less important expert $W_{n'}$ selected for compression. During fine-tuning, its effective parameter matrix $W^*_{n'}$ is computed as a weighted combination of three components: the original expert $W_{n'}$, the shared base $W_{\text{share}}$, and the expert-specific low-rank adaptation term $B_{n'} A_{n'}$. The combined parameter is given by:
\begin{equation}
    W^*_{n'} = \beta W_{n'} + (1- \beta) W_{\text{share}} + B_{n'} A_{n'},
\end{equation}
where $\beta \in [0, 1]$ is an annealing factor that is gradually decays from 1 to 0 over the course of fine-tuning. At the beginning of training, $\beta = 1$, so the model behaves identically to the original MoE. As training progresses and $\beta$ decays, the model incrementally shifts towards using the compressed representation. This progressive interpolation ensures a smooth adaptation to the shared and low-rank parameter space.
At the end of fine-tuning, $\beta = 0$, and the original expert parameters $W_{n'}$ are no longer used. They can thus be safely removed during inference, reducing the model size and achieving the desired compression. We utilize a simple yet effective decay strategy for $\beta$:
\begin{equation}
    \beta =\max \left( 1-\frac{t}{\epsilon T},0 \right) ,
\end{equation}
where $T$ is the total steps of training and $t$ is the current step. The parameter $\epsilon \in [0, 1]$ is the end ratio that controls when $\beta$ completely decays to 0. For instance, when $\epsilon=0.4$, $\beta$ reaches zero at $t=0.4T$. When $\epsilon=0$, $\beta$ is set to zero at the beginning of training, which means directly replacing the original experts with the parameter-efficient modules without any annealing period.

\section{Experiments}

\subsection{Experiment Setup}

\textbf{Datasets and evaluation.}
Following ESFT \cite{esft}, we evaluate our \MethodName{} method in two LLM customization scenarios: (1) preserving \textbf{specific ability in a domain} in which the model already demonstrates reasonable performance, and (2) adapting the model to a \textbf{narrow but unfamiliar specialized task} while achieving compression. \uline{For preservation}, we target Math, Coding, and Commonsense Reasoning domains. For the Math domain, we train on MetaMathQA \cite{metamath} and evaluate using GSM8K \cite{gsm8k}. For the Code domain, we use CodeFeedback \cite{codefeedback} for training and HumanEval \cite{humaneval} for evaluation. Evaluation protocols are consistent with those used in existing works \cite{OLMoE,ivison2023camels,wang2023far}. For the Commonsense Reasoning domain, we utilize the Cleaned Alpaca Dataset \cite{alpaca} for training. Following standard evaluation protocols \cite{alpaca, dora}, we report the averaged accuracy on eight representative commonsense reasoning tasks, such as ARC \cite{arc}, BoolQ \cite{boolq}, PIQA \cite{piqa}, and WinoGrande \cite{winogrande}. \uline{For adaptation}, we focus on intent recognition and low-resource translation. The intent recognition task comes from the BDCI-21 Smart HCI NLU Challenge.\footnote{\url{https://www.datafountain.cn/competitions/511}} The low-resource translation task utilizes the ChrEn dataset \cite{chren}, which requires translating Cherokee into English. In line with ESFT \cite{esft}, during evaluation, we compute the exact match rate between model predictions and reference answers on intent recognition task. For the translation task, we use GPT-4 to assign a score between 0 and 10 according to output quality relative to the reference.\footnote{The exact version used is \texttt{gpt-4-1106-preview}. The scores are normalized to match the scale of other reported metrics.}

\textbf{Backbone model.}
We use OLMoE-1B-7B-SFT~\cite{OLMoE} with 6.9B total and 1.3B active parameters as the pretrained model in our experiments. The model includes a fine-grained set of 64 experts for each MoE layer.

\textbf{Baselines.} We compare our proposals to a comprehensive set of baselines.
\uline{First}, we evaluate state-of-the-art expert compression methods. We adopt MC-SMoE \cite{mcsmoe} in two variants. The original version initializes by merging experts to a 30\% compression ratio and employs progressive low-rank decomposition to reach 40\% and 50\% ratios. However, this approach requires computing gradients from original experts, leading to substantial overhead. Therefore, we also test a modified version using LoRA fine-tuning to ensure a fair computational comparison. Additionally, we include HC-SMoE \cite{hcsmoe}, which employs hierarchical clustering based on expert outputs to ensure robust merging independent of routing decisions. Furthermore, we evaluate MoBE \cite{mobe}, which decomposes expert weights into shared basis matrices and expert-specific transformations.
% \textbf{Baselines.} We compare our proposals to a comprehensive set of baselines.
% \uline{First}, we evaluate state-of-the-art expert compression methods. We adopt MC-SMoE \cite{mcsmoe}, testing both its original resource-intensive version and a modified version using LoRA fine-tuning to ensure a fair comparison with our method. Additionally, we include HC-SMoE \cite{hcsmoe}, which employs hierarchical clustering based on expert outputs to ensure robust merging independent of routing decisions. Furthermore, we evaluate MoBE \cite{mobe}, which decomposes expert weights into shared basis matrices and expert-specific transformations.
% \uline{First}, we adopt MC-SMoE~\cite{mcsmoe}, which merges experts based on routing and then compresses them using gradients. As the compression is too resource-intensive, we replace it with LoRA~\cite{hu2022lora} fine-tuning after merging to ensure a similar computation complexity to our method. 
\uline{Second}, due to the lack of prior work on expert replacing, we introduce two strong baselines: ``Replace (w/o shared)'' and ``Replace (w. shared)'', which directly replace less important experts with LoRA adapters, either without adding shared base or with one shared base, respectively. Both use LoRA fine-tuning to recover performance.
\uline{Finally}, we include full fine-tuning and LoRA fine-tuning on the original model to benchmark the performance upper bound.

\textbf{Training details.} All methods use the AdamW optimizer \cite{adamw} with a batch size of 32, and a learning rate of 1e-4. For our method, we adopt a low-rank of 16 for all expert parameters and set the group size to 3. The low-rank matrices are initialized following the standard strategy \cite{hu2022lora}, which uses a random Gaussian initialization for $A$ and zero for $B$. To ensure a fair comparison, the rank for baseline methods is adjusted to maintain an equivalent number of trainable parameters. For model preservation, we adopt a training step setting of 2000. For model adaptation, training is limited to 500 steps due to smaller datasets. The optimal end ratio $\epsilon$ is determined by grid search from \{0.1, 0.2, 0.3, 0.4, 0.5\}.

\subsection{Main Results}

% main_results
\begin{table*}[ht]
\small
% 1. 调整行高倍率，并缩小列间距以容纳新列
\renewcommand{\arraystretch}{1.1} 
\renewcommand\tabcolsep{4.0pt}
\centering
% 2. 移除额外间距设置
\setlength{\extrarowheight}{0pt}
\setlength{\aboverulesep}{0pt}
\setlength{\belowrulesep}{0pt}
\caption{Main performance comparison across methods and tasks at different compression ratios. `*' indicates the full method as described in the original paper. ``\# Params'' is the number of trainable parameters. Best results are shown in \textbf{bold} and second-best results are \underline{underlined}. Our method \MethodName{} consistently achieves good performance among all tasks under different compression settings.}
\label{tb:main_results}
\begin{tabular}{c|c|c|cccccc}
\toprule
\hline 
    &                       &                       & \textbf{Math} & \textbf{Code} & \textbf{Commonsense} & \multicolumn{2}{c}{\textbf{Specialized Tasks}} &            \\ 
    \cmidrule(lr){4-4} \cmidrule(lr){5-5} \cmidrule(lr){6-6} \cmidrule(lr){7-8}
\multirow{-2}{*}{\textbf{Ratio}}  & 
  \multirow{-2}{*}{\textbf{Methods}} &
  \multirow{-2}{*}{\textbf{\# Params}} &
  GSM8K &
  HumanEval &
  8 Sub-Tasks &
  Intent &
  Translation &
  \multirow{-2}{*}{Average} 
  \\ \hline 
0\% & Original             & -     & 49.3          & 55.0          & 60.6          & 9.8                   & 13.6                    & 37.7       \\
0\% & Full FT              & 6.92B & 61.2          & 49.7          & 54.2          & 79.2                  & 37.1                    & 56.3       \\ 
0\% & LoRA                 & 0.45B & 59.8          & 54.2          & 61.4          & 73.0                  & 29.2                    & 55.5      
\\ \hline 
    & MoBE                 & 0.45B & 52.9          & 41.3          & \textbf{58.3} & 69.4                  & 30.3                    & 50.4       \\
    & HC-SMoE              & 0.45B & 40.6          & 41.7          & 55.5          & 64.0                  & \textbf{32.4}           & 46.8       \\
    & MC-SMoE              & 0.45B & \underline{59.0} & 51.7       & 55.1          & 70.2                  & \underline{30.9}        & \underline{53.4} \\
    & Replace (w/o shared) & 0.45B & 57.2          & \underline{53.8} & 55.6       & 69.4                  & 27.0                    & 52.6       \\
    & Replace (w. shared)  & 0.45B & 57.8          & 52.0          & \underline{56.7} & \underline{71.4}    & 23.6                    & 52.3       \\
\multirow{-6}{*}{30\%} &
  \cellcolor[HTML]{EFEFEF}\textbf{\MethodName{}} &
  \cellcolor[HTML]{EFEFEF}0.45B &
  \cellcolor[HTML]{EFEFEF}\textbf{59.7} &
  \cellcolor[HTML]{EFEFEF}\textbf{55.7} &
  \cellcolor[HTML]{EFEFEF}56.3 &
  \cellcolor[HTML]{EFEFEF}\textbf{74.6} &
  \cellcolor[HTML]{EFEFEF}30.4 &
  \cellcolor[HTML]{EFEFEF}\textbf{55.3}
  \\ \hline 
    & MC-SMoE* & 1.65B & 57.0          & \underline{50.6} & 40.3         & \textbf{74.5}         & 28.4                    & \underline{50.2} \\
    & MoBE                 & 0.45B & 37.2          & 28.6          & \textbf{54.0} & 65.4                  & \underline{29.9}        & 43.0       \\
    & HC-SMoE              & 0.45B & 31.2          & 33.5          & 51.5          & 58.8                  & 16.9                    & 38.4       \\
    & MC-SMoE              & 0.45B & 56.0          & 48.4          & 50.4          & 70.8                  & 10.8                    & 47.3       \\
    & Replace (w/o shared) & 0.45B & 56.1          & \textbf{51.3} & 52.5          & 65.8                  & 24.7                    & 50.1       \\
    & Replace (w. shared)  & 0.45B & \underline{57.7} & 50.1       & \underline{53.9} & 59.0               & \underline{25.9}        & 49.3       \\
\multirow{-7}{*}{40\%} &
  \cellcolor[HTML]{EFEFEF}\textbf{\MethodName{}} &
  \cellcolor[HTML]{EFEFEF}0.45B &
  \cellcolor[HTML]{EFEFEF}\textbf{57.9} &
  \cellcolor[HTML]{EFEFEF}\textbf{51.3} &
  \cellcolor[HTML]{EFEFEF}53.2 &
  \cellcolor[HTML]{EFEFEF}\underline{72.6} &
  \cellcolor[HTML]{EFEFEF}\textbf{30.2} &
  \cellcolor[HTML]{EFEFEF}\textbf{53.0}
  \\ \hline 
    & MC-SMoE* & 1.65B & 50.8          & 43.8          & 40.0          & \underline{73.4}      & \underline{18.5}        & \underline{45.3} \\
    & MoBE                 & 0.45B & 18.7          & 9.7           & 46.4          & 53.0                  & \textbf{21.6}           & 29.9       \\
    & HC-SMoE              & 0.45B & 26.2          & 26.5          & 47.8          & 58.0                  & 11.6                    & 34.0       \\
    & MC-SMoE              & 0.45B & 50.1          & 43.1          & 44.8          & 67.2                  & 7.2                     & 42.5       \\
    & Replace (w/o shared) & 0.45B & 49.5          & \underline{46.3} & 47.1       & 65.4                  & 13.0                    & 44.3       \\
    & Replace (w. shared)  & 0.45B & \underline{51.4} & 44.5       & \textbf{49.3} & 56.2                  & 6.7                     & 41.6       \\
\multirow{-7}{*}{50\%} &
  \cellcolor[HTML]{EFEFEF}\textbf{\MethodName{}} &
  \cellcolor[HTML]{EFEFEF}0.45B &
  \cellcolor[HTML]{EFEFEF}\textbf{56.1} &
  \cellcolor[HTML]{EFEFEF}\textbf{48.4} &
  \cellcolor[HTML]{EFEFEF}\underline{48.1} &
  \cellcolor[HTML]{EFEFEF}\textbf{74.2} &
  \cellcolor[HTML]{EFEFEF}13.5          &
  \cellcolor[HTML]{EFEFEF}\textbf{48.1}
  \\ \hline
\bottomrule
\end{tabular}
% \vspace{-0.2cm}
\end{table*}

\cref{tb:main_results} reports the main results on five tasks at three compression ratios. Our method consistently achieves the best or second-best performance across various compression ratios.
Remarkably, at a 30\% compression ratio, our method performs comparably to LoRA, even surpassing it on some tasks. This suggests that replacing extremely unimportant experts with parameter-efficient modules even benefits downstream task performance.
At a 50\% compression ratio with the same training budget, our method significantly outperforms existing methods, achieving average performance improvements of 5.6\% and 3.8\% over existing methods and the directly replacing baseline, respectively. Besides, even with over three times the trainable parameters, MC-SMoE* still lags behind our method by 2.8\%. This demonstrates that our method strikes a superior balance among compression efficiency, training efficiency, and model performance.

For model preservation tasks, our method maintains robust performance even under high compression ratios. Notably, on the Math task, our approach preserves 94\% of LoRA's performance while reducing the model parameters by 50\%. This demonstrates that our method can effectively preserve the existing ability of the model.
For model adaptation tasks, although performance fluctuates across different specialized tasks and compression ratios, our method outperforms alternative approaches. This indicates that our technique successfully enables model adaptation to an unfamiliar downstream task while simultaneously achieving model compression.
% For model adaptation tasks, although performance fluctuates across different specialized tasks and compression ratios, our method outperforms alternative approaches. This indicates that our technique successfully enables model adaptation to an unfamiliar downstream task while simultaneously achieving model compression. 

Furthermore, the ``Replace (w/o shared)'' method achieves performance comparable to MC-SMoE at a 30\% ratio and surpasses it at 40\%. Notably, even compared to MC-SMoE*, it maintains comparable performance at both 40\% and 50\% ratios. This shows that directly replacing less important experts is a strong baseline for compressing MoE models.
% Furthermore, at a 30\% compression ratio, the ``Replace (w/o shared)'' method achieves performance comparable to MC-SMoE, and at a 40\% compression ratio, it surpasses MC-SMoE. This indicates that directly replacing less important experts is a strong baseline for compressing MoE models. Although the``Replace (w. shared)'' method outperforms ``Replace (w/o shared)'' on the Math task, its overall performance remains sub-optimal. In contrast, our method consistently outperforms these approaches, highlighting the benefits of our multiple shared experts strategy and the annealed expert replacement mechanism in enhancing the stability of expert compression.

\subsection{Ablation Study}

\begin{table}[ht]
\small
% 1. 调整行高倍率，列间距稍微调大（因为只有4列，4pt会太挤）
\renewcommand{\arraystretch}{1.1}
\renewcommand\tabcolsep{8.0pt}
\centering

% 2. 移除额外间距设置，保持与主表一致的紧凑风格
\setlength{\extrarowheight}{0pt}
\setlength{\aboverulesep}{0pt}
\setlength{\belowrulesep}{0pt}

\caption{Comparison of different experts selection methods for \MethodName{} on the Math and Code tasks. Our adaptive thresholding method can select the most appropriate experts for replacing, adapting to different compression ratios.}
\label{tb:selection method}

\begin{tabular}{c|c|cc}
\toprule
\hline
\multirow{2}{*}{\textbf{Ratio}} & \multirow{2}{*}{\textbf{Selection Methods}} & \textbf{Math} & \textbf{Code} \\ 
\cmidrule(lr){3-3} \cmidrule(lr){4-4}
 & & GSM8K & HumanEval \\ \hline
\multirow{3}{*}{30\%} & Uniform & 58.1 & 55.1 \\
 & Average & \underline{59.4} & \underline{55.5} \\
 & \cellcolor[HTML]{EFEFEF}\textbf{Adaptive} & \cellcolor[HTML]{EFEFEF}\textbf{59.7} & \cellcolor[HTML]{EFEFEF}\textbf{55.7} \\ \hline
\multirow{3}{*}{50\%} & Uniform & \textbf{57.1} & \underline{47.9} \\
 & Average & 54.6 & 43.0 \\
 & \cellcolor[HTML]{EFEFEF}\textbf{Adaptive} & \cellcolor[HTML]{EFEFEF}\underline{56.1} & \cellcolor[HTML]{EFEFEF}\textbf{48.4} \\ \hline
\bottomrule
\end{tabular}
\end{table}

\begin{table}[!ht]
\small
% 1. 调整行高倍率
\renewcommand{\arraystretch}{1.1}
% 2. 调整列间距（因为只有3列，稍微调大至12pt以保持平衡）
\renewcommand\tabcolsep{12.0pt}
\centering
% 3. 移除额外间距设置，保持紧凑风格
\setlength{\extrarowheight}{0pt}
\setlength{\aboverulesep}{0pt}
\setlength{\belowrulesep}{0pt}
\caption{Comparison of expert grouping strategies for OLMoE on the Math task at different compression ratios.}
\label{tab:grouping_comparison}
\begin{tabular}{c|cc}
\toprule
\hline
\textbf{Ratio} & \textbf{K-means} & \textbf{Dominant (Ours)} \\ \hline
30\% & \textbf{60.0} & 59.7 \\
40\% & 56.1 & \textbf{57.9} \\
50\% & 54.9 & \textbf{56.1} \\ \hline
\bottomrule
\end{tabular}
\end{table}

% \begin{figure*}[!ht]
%     \centering
%       \includegraphics[width=0.9\textwidth]{src/group_size.pdf}
%     \caption{Comparison of different group sizes at different compression ratios on the Math task.}
%     \label{fig:group_size}
%     \vspace{-0.4cm}
% \end{figure*}

\begin{figure*}[!ht]
    \centering
      \includegraphics[width=0.75\textwidth]{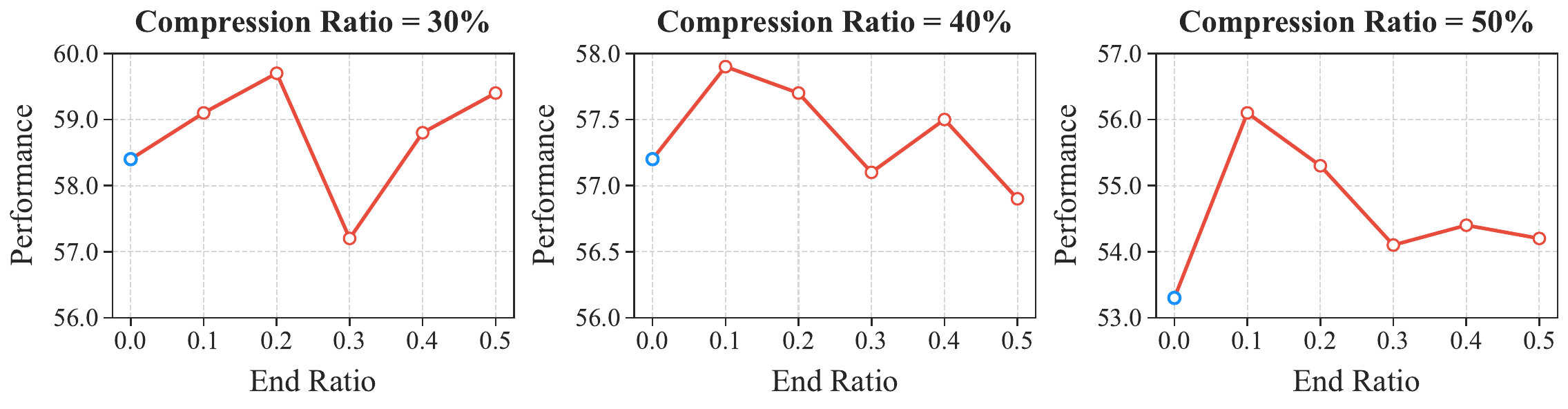}
    % \vspace{-8pt}
    \caption{Comparison of different end ratios for \MethodName{} on the Math task. Directly replacing (blue points) is consistently sub-optimal, which shows the effectiveness of our annealed expert replacement strategy (red points).}
    % \vspace{-8pt}
    \label{fig:end ratio}
\end{figure*}

\begin{figure*}[!ht]
    \centering
      \includegraphics[width=0.75\textwidth]{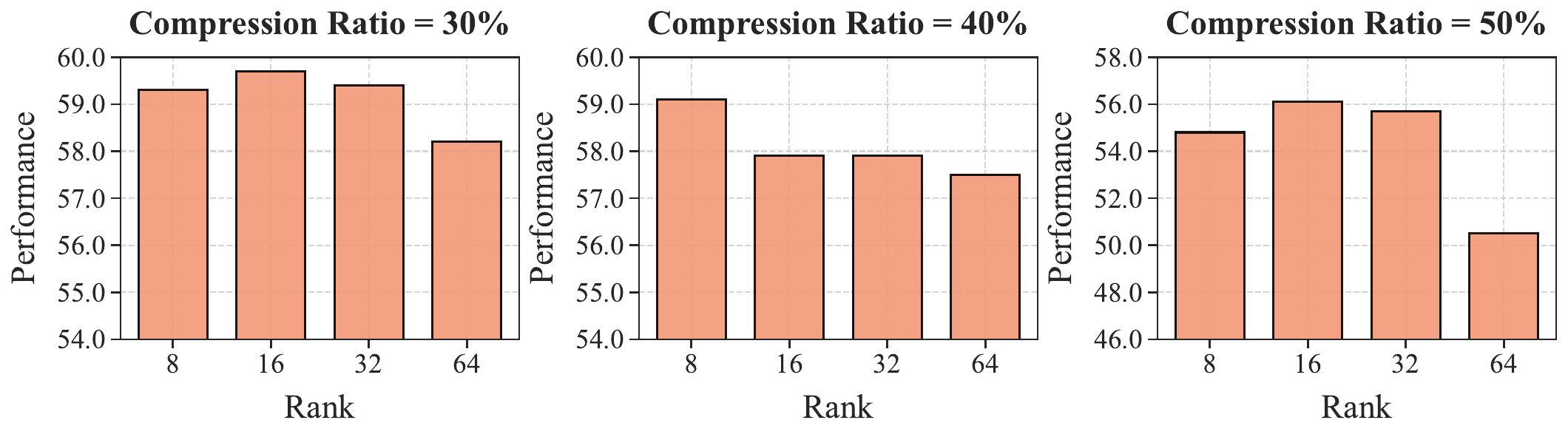}
    % \vspace{-8pt}
    \caption{Comparison of different ranks at different compression ratios on the Math task.}
    % \vspace{-8pt}
    \label{fig:lora rank}
\end{figure*}

% \begin{table}[!ht]
%     \centering
%     \caption{Comparison of expert grouping strategies for OLMoE at different compression ratios.}
%     \label{tab:grouping_comparison}
%     \begin{tabular}{c|cc}
%         \toprule
%         \textbf{Ratio} & \textbf{K-means} & \textbf{Dominant (Ours)} \\
%         \midrule
%         30\% & \textbf{60.0} & 59.7 \\
%         40\% & 56.1 & \textbf{57.9} \\
%         50\% & 54.9 & \textbf{56.1} \\
%         \bottomrule
%     \end{tabular}
% \end{table}

\textbf{Impact of expert selection scheme.}
To assess the effectiveness of our adaptive thresholding approach, we conduct an ablation study on different expert selection methods: ``Uniform'' (constant threshold per layer), ``Average'' (same number of selected experts per layer), and ``Adaptive'' (ours). Experimental results are produced on the Math and Code tasks, as shown in \cref{tb:selection method}. 
At a low compression ratio, our method performs comparably to the ``Average'' selection method but outperforms the ``Uniform'' selection method. At a high compression ratio, our method performs similarly to the ``Uniform'' but significantly outperforms the ``Average''. This indicates that our method can select the most appropriate experts for replacing, adapting to different compression ratios effectively.

\textbf{Impact of expert grouping strategy.}
To explore different ways of grouping experts, we compare our method with the K-means clustering strategy, which is adapted from HC-SMoE \cite{hcsmoe}. 
As shown in \cref{tab:grouping_comparison}, both strategies perform similarly at a mild 30\% ratio. However, as the compression becomes more aggressive, our method demonstrates a clear advantage. At the 40\% and 50\% ratios, our approach significantly outperforms K-means.
% achieving scores of 57.9\% and 56.1\% compared to 56.1\% and 54.9\%, respectively.
We attribute this performance gap to the distinct mechanisms of group formation. Our method focuses on preserving the "dominant" experts and grouping similar experts around them. In contrast, K-means computes average centroids for all experts. When the compression is high,  K-means risks inadvertently merging critical experts into a general group, leading to the irreversible loss of expert-specific knowledge. This suggests that explicitly preserving dominant experts is crucial to maintaining performance under high compression.

% \textbf{Impact of group size.}
% We explore the impact of incrementally increasing the group size of \MethodName{} from 1 to 4. \cref{fig:group_size} illustrates performance across various group sizes and compression ratios on the Math task.
% At a mild 30\% compression ratio, the performance of all group sizes vary by less than 0.9\%, showing that multiple shared experts has little effect when ample parameters remain.

% In contrast, as compression becomes more aggressive, the benefit of larger group sizes becomes pronounced. In particular, at a 50\% compression ratio, performance improves with increasing group size. This trend suggests that under constrained parameter budgets, constructing multiple shared experts by grouping is crucial for preserving the model’s knowledge and capabilities.

% \begin{figure*}
%     \centering
%     \includegraphics[width=\textwidth]{src/main.pdf}
%     \caption{Overview of the proposed \MethodName{} framework. (\textit{Left}) A standard MoE layer. (\textit{Right}) The \MethodName{} workflow, comprising three key steps: (1) expert scoring and selecting the low-scoring experts as compression candidates, (2) candidate grouping, and (3) replacement of each group with a shared expert augmented by lightweight expert-specific adaptation parameters.}
%     \label{fig:main}
%     % \vspace{-0.4cm}
% \end{figure*}

\textbf{Impact of decay ratio.}
\cref{fig:end ratio} illustrates how the end ratio hyperparameter affects model performance. Directly replacing (blue points) is consistently sub-optimal, which shows the effectiveness of our annealed expert replacement strategy. Furthermore, we find that that lower end ratios, specifically 0.1 or 0.2, can often yield the best results.

\textbf{Impact of adaptation rank.}
To investigate the impact of adaptation rank, we conduct an ablation study across various ranks using the Math task, with results presented in \cref{fig:lora rank}. 
Overall, we observe that lower ranks generally yield superior performance. 
% For a 30\% compression ratio, performance peaks at rank 16, with a slight drop beyond that. At 40\% compression, the best performance is achieved at rank 8, and increasing the rank leads to a consistent decline. A similar trend is observed at 50\% compression, where rank 16 offers the highest performance, and further increases result in degradation. 
Specifically, at a 30\% compression ratio, performance peaks at rank 16, with a slight drop at higher ranks. Similarly, at 40\% and 50\% compression ratios, the best performance is achieved at relatively low ranks, while increasing the rank leads to performance degradation.
These findings suggest that smaller ranks are generally more effective at higher compression levels.

\section{Conclusion}

In this paper, we introduce a novel expert compression paradigm termed expert replacing. Our empirical findings demonstrate that even a simple baseline of this paradigm yields promising results. Building on this foundation, we propose \MethodName{}, a framework that enhances the paradigm by introducing adaptive expert selection, hierarchical expert construction, and an annealed recovery strategy.
Experimental results across diverse tasks confirm the effectiveness of our approach. \MethodName{} achieves performance comparable to LoRA fine-tuning at a 30\% compression ratio and significantly outperforms existing state-of-the-art methods at a more aggressive 50\% ratio. These results indicate that our method successfully reduces the memory footprint of MoE models with minimal training overhead while avoiding substantial performance degradation. This work not only provides a practical solution for compressing MoE models but also opens up a broader horizon for further research on the expert replacing paradigm. Future work could explore advanced initialization methods and adaptive rank allocation strategies to further unlock the potential of this paradigm.

% In this work, we proposed \MethodName{}, a framework for compressing redundant experts in MoE models. Our approach identifies less important experts, replaces them with a smaller set of shared experts augmented with task-specific low-rank adaptation modules, and employs a progressive annealing process to ensure a smooth transition during fine-tuning. Experimental results across diverse tasks show that \MethodName{} achieves competitive performance to LoRA-based fine-tuning at a 30\% compression ratio, while at a more aggressive 50\% compression ratio, it outperforms existing methods by an average of 5.3\%. Our results suggest that \MethodName{} is an effective solution for reducing the memory footprint of MoE models without incurring significant performance degradation, thereby facilitating more efficient and scalable deployment of MoE models.

% In the unusual situation where you want a paper to appear in the
% references without citing it in the main text, use \nocite

\bibliography{ref}

@inproceedings{esft,
  author       = {Zihan Wang and
                  Deli Chen and
                  Damai Dai and
                  Runxin Xu and
                  Zhuoshu Li and
                  Yu Wu},
  title        = {Let the Expert Stick to His Last: Expert-Specialized Fine-Tuning for
                  Sparse Architectural Large Language Models},
  booktitle    = {{EMNLP}},
  pages        = {784--801},
  publisher    = {Association for Computational Linguistics},
  year         = {2024}
}

@article{song2024layer,
  title={Layer Importance and Hallucination Analysis in Large Language Models via Enhanced Activation Variance-Sparsity},
  author={Song, Zichen and Huang, Sitan and Wu, Yuxin and Kang, Zhongfeng},
  journal={arXiv preprint arXiv:2411.10069},
  year={2024}
}

@article{hu2022lora,
  title={Lora: Low-rank adaptation of large language models.},
  author={Hu, Edward J and Shen, Yelong and Wallis, Phillip and Allen-Zhu, Zeyuan and Li, Yuanzhi and Wang, Shean and Wang, Lu and Chen, Weizhu and others},
  journal={ICLR},
  volume={1},
  number={2},
  pages={3},
  year={2022}
}

@article{touvron2023llama,
  title={Llama 2: Open foundation and fine-tuned chat models},
  author={Touvron, Hugo and Martin, Louis and Stone, Kevin and Albert, Peter and Almahairi, Amjad and Babaei, Yasmine and Bashlykov, Nikolay and Batra, Soumya and Bhargava, Prajjwal and Bhosale, Shruti and others},
  journal={arXiv preprint arXiv:2307.09288},
  year={2023}
}

@article{grattafiori2024llama,
  title={The llama 3 herd of models},
  author={Grattafiori, Aaron and Dubey, Abhimanyu and Jauhri, Abhinav and Pandey, Abhinav and Kadian, Abhishek and Al-Dahle, Ahmad and Letman, Aiesha and Mathur, Akhil and Schelten, Alan and Vaughan, Alex and others},
  journal={arXiv preprint arXiv:2407.21783},
  year={2024}
}

@article{lepikhin2020gshard,
  title={Gshard: Scaling giant models with conditional computation and automatic sharding},
  author={Lepikhin, Dmitry and Lee, HyoukJoong and Xu, Yuanzhong and Chen, Dehao and Firat, Orhan and Huang, Yanping and Krikun, Maxim and Shazeer, Noam and Chen, Zhifeng},
  journal={arXiv preprint arXiv:2006.16668},
  year={2020}
}

@article{dai2022stablemoe,
  title={Stablemoe: Stable routing strategy for mixture of experts},
  author={Dai, Damai and Dong, Li and Ma, Shuming and Zheng, Bo and Sui, Zhifang and Chang, Baobao and Wei, Furu},
  journal={arXiv preprint arXiv:2204.08396},
  year={2022}
}

@article{shen2024jetmoe,
  title={Jetmoe: Reaching llama2 performance with 0.1 m dollars},
  author={Shen, Yikang and Guo, Zhen and Cai, Tianle and Qin, Zengyi},
  journal={arXiv preprint arXiv:2404.07413},
  year={2024}
}

@article{jiang2024mixtral,
  title={Mixtral of experts},
  author={Jiang, Albert Q and Sablayrolles, Alexandre and Roux, Antoine and Mensch, Arthur and Savary, Blanche and Bamford, Chris and Chaplot, Devendra Singh and Casas, Diego de las and Hanna, Emma Bou and Bressand, Florian and others},
  journal={arXiv preprint arXiv:2401.04088},
  year={2024}
}

@inproceedings{deepseekmoe,
  author       = {Damai Dai and
                  Chengqi Deng and
                  Chenggang Zhao and
                  R. X. Xu and
                  Huazuo Gao and
                  Deli Chen and
                  Jiashi Li and
                  Wangding Zeng and
                  Xingkai Yu and
                  Y. Wu and
                  Zhenda Xie and
                  Y. K. Li and
                  Panpan Huang and
                  Fuli Luo and
                  Chong Ruan and
                  Zhifang Sui and
                  Wenfeng Liang},
  title        = {DeepSeekMoE: Towards Ultimate Expert Specialization in Mixture-of-Experts Language Models},
  booktitle    = {{ACL} {(1)}},
  pages        = {1280--1297},
  publisher    = {Association for Computational Linguistics},
  year         = {2024}
}

@article{deepseekv2,
  title={Deepseek-v2: A strong, economical, and efficient mixture-of-experts language model},
  author={Liu, Aixin and Feng, Bei and Wang, Bin and Wang, Bingxuan and Liu, Bo and Zhao, Chenggang and Dengr, Chengqi and Ruan, Chong and Dai, Damai and Guo, Daya and others},
  journal={arXiv preprint arXiv:2405.04434},
  year={2024}
}

@inproceedings{OLMoE,
  author       = {Niklas Muennighoff and
                  Luca Soldaini and
                  Dirk Groeneveld and
                  Kyle Lo and
                  Jacob Morrison and
                  Sewon Min and
                  Weijia Shi and
                  Evan Pete Walsh and
                  Oyvind Tafjord and
                  Nathan Lambert and
                  Yuling Gu and
                  Shane Arora and
                  Akshita Bhagia and
                  Dustin Schwenk and
                  David Wadden and
                  Alexander Wettig and
                  Binyuan Hui and
                  Tim Dettmers and
                  Douwe Kiela and
                  Ali Farhadi and
                  et al.},
  title        = {OLMoE: Open Mixture-of-Experts Language Models},
  booktitle    = {{ICLR}},
  publisher    = {OpenReview.net},
  year         = {2025}
}

@article{qwen2,
    title   = {Qwen2 Technical Report}, 
    author  = {An Yang and Baosong Yang and Binyuan Hui and Bo Zheng and Bowen Yu and Chang Zhou and Chengpeng Li and Chengyuan Li and Dayiheng Liu and Fei Huang and Guanting Dong and Haoran Wei and Huan Lin and Jialong Tang and Jialin Wang and Jian Yang and Jianhong Tu and Jianwei Zhang and Jianxin Ma and Jin Xu and Jingren Zhou and Jinze Bai and Jinzheng He and Junyang Lin and Kai Dang and Keming Lu and Keqin Chen and Kexin Yang and Mei Li and Mingfeng Xue and Na Ni and Pei Zhang and Peng Wang and Ru Peng and Rui Men and Ruize Gao and Runji Lin and Shijie Wang and Shuai Bai and Sinan Tan and Tianhang Zhu and Tianhao Li and Tianyu Liu and Wenbin Ge and Xiaodong Deng and Xiaohuan Zhou and Xingzhang Ren and Xinyu Zhang and Xipin Wei and Xuancheng Ren and Yang Fan and Yang Yao and Yichang Zhang and Yu Wan and Yunfei Chu and Yuqiong Liu and Zeyu Cui and Zhenru Zhang and Zhihao Fan},
    journal = {arXiv preprint arXiv:2407.10671},
    year    = {2024}
}

@article{qwen2.5,
    title   = {Qwen2.5 Technical Report}, 
    author  = {An Yang and Baosong Yang and Beichen Zhang and Binyuan Hui and Bo Zheng and Bowen Yu and Chengyuan Li and Dayiheng Liu and Fei Huang and Haoran Wei and Huan Lin and Jian Yang and Jianhong Tu and Jianwei Zhang and Jianxin Yang and Jiaxi Yang and Jingren Zhou and Junyang Lin and Kai Dang and Keming Lu and Keqin Bao and Kexin Yang and Le Yu and Mei Li and Mingfeng Xue and Pei Zhang and Qin Zhu and Rui Men and Runji Lin and Tianhao Li and Tingyu Xia and Xingzhang Ren and Xuancheng Ren and Yang Fan and Yang Su and Yichang Zhang and Yu Wan and Yuqiong Liu and Zeyu Cui and Zhenru Zhang and Zihan Qiu},
    journal = {arXiv preprint arXiv:2412.15115},
    year    = {2024}
}

@inproceedings{metamath,
  author       = {Longhui Yu and
                  Weisen Jiang and
                  Han Shi and
                  Jincheng Yu and
                  Zhengying Liu and
                  Yu Zhang and
                  James T. Kwok and
                  Zhenguo Li and
                  Adrian Weller and
                  Weiyang Liu},
  title        = {MetaMath: Bootstrap Your Own Mathematical Questions for Large Language
                  Models},
  booktitle    = {{ICLR}},
  publisher    = {OpenReview.net},
  year         = {2024}
}

@article{gsm8k,
  title={Training verifiers to solve math word problems},
  author={Cobbe, Karl and Kosaraju, Vineet and Bavarian, Mohammad and Chen, Mark and Jun, Heewoo and Kaiser, Lukasz and Plappert, Matthias and Tworek, Jerry and Hilton, Jacob and Nakano, Reiichiro and others},
  journal={arXiv preprint arXiv:2110.14168},
  year={2021}
}

@inproceedings{codefeedback,
  author       = {Tianyu Zheng and
                  Ge Zhang and
                  Tianhao Shen and
                  Xueling Liu and
                  Bill Yuchen Lin and
                  Jie Fu and
                  Wenhu Chen and
                  Xiang Yue},
  title        = {OpenCodeInterpreter: Integrating Code Generation with Execution and
                  Refinement},
  booktitle    = {{ACL} (Findings)},
  pages        = {12834--12859},
  publisher    = {Association for Computational Linguistics},
  year         = {2024}
}

@article{humaneval,
  author       = {Mark Chen and
                  Jerry Tworek and
                  Heewoo Jun and
                  Qiming Yuan and
                  Henrique Pond{\'{e}} de Oliveira Pinto and
                  Jared Kaplan and
                  Harri Edwards and
                  Yuri Burda and
                  Nicholas Joseph and
                  Greg Brockman and
                  Alex Ray and
                  Raul Puri and
                  Gretchen Krueger and
                  Michael Petrov and
                  Heidy Khlaaf and
                  Girish Sastry and
                  Pamela Mishkin and
                  Brooke Chan and
                  Scott Gray and
                  Nick Ryder and
                  Mikhail Pavlov and
                  Alethea Power and
                  Lukasz Kaiser and
                  Mohammad Bavarian and
                  Clemens Winter and
                  Philippe Tillet and
                  Felipe Petroski Such and
                  Dave Cummings and
                  Matthias Plappert and
                  Fotios Chantzis and
                  Elizabeth Barnes and
                  Ariel Herbert{-}Voss and
                  William Hebgen Guss and
                  Alex Nichol and
                  Alex Paino and
                  Nikolas Tezak and
                  Jie Tang and
                  Igor Babuschkin and
                  Suchir Balaji and
                  Shantanu Jain and
                  William Saunders and
                  Christopher Hesse and
                  Andrew N. Carr and
                  Jan Leike and
                  Joshua Achiam and
                  Vedant Misra and
                  Evan Morikawa and
                  Alec Radford and
                  Matthew Knight and
                  Miles Brundage and
                  Mira Murati and
                  Katie Mayer and
                  Peter Welinder and
                  Bob McGrew and
                  Dario Amodei and
                  Sam McCandlish and
                  Ilya Sutskever and
                  Wojciech Zaremba},
  title        = {Evaluating Large Language Models Trained on Code},
  journal      = {CoRR},
  volume       = {abs/2107.03374},
  year         = {2021}
}

@misc{ivison2023camels,
      title={Camels in a Changing Climate: Enhancing LM Adaptation with Tulu 2}, 
      author={Hamish Ivison and Yizhong Wang and Valentina Pyatkin and Nathan Lambert and Matthew Peters and Pradeep Dasigi and Joel Jang and David Wadden and Noah A. Smith and Iz Beltagy and Hannaneh Hajishirzi},
      year={2023},
      eprint={2311.10702},
      archivePrefix={arXiv},
      primaryClass={cs.CL}
}

@misc{wang2023far,
   title={How Far Can Camels Go? Exploring the State of Instruction Tuning on Open Resources}, 
   author={Yizhong Wang and Hamish Ivison and Pradeep Dasigi and Jack Hessel and Tushar Khot and Khyathi Raghavi Chandu and David Wadden and Kelsey MacMillan and Noah A. Smith and Iz Beltagy and Hannaneh Hajishirzi},
   year={2023},
   eprint={2306.04751},
   archivePrefix={arXiv},
   primaryClass={cs.CL}
}

@inproceedings{chren,
  author       = {Shiyue Zhang and
                  Benjamin Frey and
                  Mohit Bansal},
  title        = {ChrEn: Cherokee-English Machine Translation for Endangered Language
                  Revitalization},
  booktitle    = {{EMNLP} {(1)}},
  pages        = {577--595},
  publisher    = {Association for Computational Linguistics},
  year         = {2020}
}

@inproceedings{mcsmoe,
  author       = {Pingzhi Li and
                  Zhenyu Zhang and
                  Prateek Yadav and
                  Yi{-}Lin Sung and
                  Yu Cheng and
                  Mohit Bansal and
                  Tianlong Chen},
  title        = {Merge, Then Compress: Demystify Efficient SMoE with Hints from Its
                  Routing Policy},
  booktitle    = {{ICLR}},
  publisher    = {OpenReview.net},
  year         = {2024}
}

@inproceedings{Kim2024ES,
  author       = {Yechan Kim and
                  Hwijoon Lim and
                  Dongsu Han},
  title        = {Scaling Beyond the GPU Memory Limit for Large Mixture-of-Experts
                  Model Training},
  booktitle    = {{ICML}},
  publisher    = {OpenReview.net},
  year         = {2024}
}

@article{Eliseev2023FastInference,
  author  = {Artyom Eliseev and Denis Mazur},
  title   = {Fast Inference of Mixture-of-Experts Language Models with Offloading},
  journal = {arXiv preprint arXiv:2312.17238},
  year    = {2023}
}

@article{Yu2025fMoE,
  author  = {Hanfei Yu and Xingqi Cui and Hong Zhang and Hao Wang},
  title   = {{fMoE}: Fine-Grained Expert Offloading for Large Mixture-of-Experts Serving},
  journal = {arXiv preprint arXiv:2502.05370},
  year    = {2025}
}

@article{Liu2024EEP,
  author  = {Enshu Liu and Junyi Zhu and Zinan Lin and Xuefei Ning and Matthew B. Blaschko and Shengen Yan and Guohao Dai and Huazhong Yang and Yu Wang},
  title   = {Efficient Expert Pruning for Sparse Mixture-of-Experts Language Models: Enhancing Performance and Reducing Inference Costs},
  journal = {arXiv preprint arXiv:2407.00945},
  year    = {2024}
}

@article{Lu2024NAEE,
  author  = {Xudong Lu and Qi Liu and Yuhui Xu and Aojun Zhou and Siyuan Huang and Bo Zhang and Junchi Yan and Hongsheng Li},
  title   = {Not All Experts are Equal: Efficient Expert Pruning and Skipping for Mixture-of-Experts Large Language Models},
  journal = {arXiv preprint arXiv:2402.14800},
  year    = {2024}
}

@article{yang2024moe_i^2,
  title={MoE-I2: Compressing Mixture of Experts Models through Inter-Expert Pruning and Intra-Expert Low-Rank Decomposition},
  author={Yang, Cheng and Sui, Yang and Xiao, Jinqi and Huang, Lingyi and Gong, Yu and Duan, Yuanlin and Jia, Wenqi and Yin, Miao and Cheng, Yu and Yuan, Bo},
  journal={arXiv preprint arXiv:2411.01016},
  year={2024}
}

@article{xie2024moepruner,
  title={MoE-Pruner: Pruning Mixture-of-Experts Large Language Model using the Hints from Its Router},
  author={Xie, Yanyue and Zhang, Zhi and Zhou, Ding and Xie, Cong and Song, Ziang and Liu, Xin and Wang, Yanzhi and Lin, Xue and Xu, An},
  journal={arXiv preprint arXiv:2410.12013},
  year={2024}
}

@article{Li2024Quant,
  author  = {Pingzhi Li and Xiaolong Jin and Yu Cheng and Tianlong Chen},
  title   = {Examining Post-Training Quantization for Mixture-of-Experts: A Benchmark},
  journal = {arXiv preprint arXiv:2406.08155},
  year    = {2024}
}

@article{Huang2024MCMoE,
  author  = {Wei Huang and Yue Liao and Jianhui Liu and Ruifei He and Haoru Tan and Shiming Zhang and Hongsheng Li and Si Liu and Xiaojuan Qi},
  title   = {MC-MoE: Mixture Compressor for Mixture-of-Experts LLMs Gains More},
  journal = {arXiv preprint arXiv:2410.06270},
  year    = {2024}
}

@article{Kim2025Every,
  author  = {Gyeongman Kim and Gyouk Chu and Eunho Yang},
  title   = {Every Expert Matters: Towards Effective Knowledge Distillation for Mixture-of-Experts Language Models},
  journal = {arXiv preprint arXiv:2502.12947},
  year    = {2025}
}

@inproceedings{xu2024distillation,
  title={Sparse Mixture of Experts Language Models Excel in Knowledge Distillation},
  author={Xu, Haiyang and Liu, Haoxiang and Gong, Wei and Deng, Xianjun and Wang, Hai},
  booktitle={CCF International Conference on Natural Language Processing and Chinese Computing},
  pages={80--91},
  year={2024},
  organization={Springer}
}

@article{han2024peft_survey,
  title={Parameter-efficient fine-tuning for large models: A comprehensive survey},
  author={Han, Zeyu and Gao, Chao and Liu, Jinyang and Zhang, Jeff and Zhang, Sai Qian},
  journal={arXiv preprint arXiv:2403.14608},
  year={2024}
}

@article{xu2023peft_survey,
  title={Parameter-efficient fine-tuning methods for pretrained language models: A critical review and assessment},
  author={Xu, Lingling and Xie, Haoran and Qin, Si-Zhao Joe and Tao, Xiaohui and Wang, Fu Lee},
  journal={arXiv preprint arXiv:2312.12148},
  year={2023}
}

@article{wang2022adamix,
  title={Adamix: Mixture-of-adapter for parameter-efficient tuning of large language models},
  author={Wang, Yaqing and Mukherjee, Subhabrata and Liu, Xiaodong and Gao, Jing and Awadallah, Ahmed Hassan and Gao, Jianfeng},
  journal={arXiv preprint arXiv:2205.12410},
  volume={1},
  number={2},
  pages={4},
  year={2022}
}

@inproceedings{condiadapter,
  author       = {Tao Lei and
                  Junwen Bai and
                  Siddhartha Brahma and
                  Joshua Ainslie and
                  Kenton Lee and
                  Yanqi Zhou and
                  Nan Du and
                  Vincent Y. Zhao and
                  Yuexin Wu and
                  Bo Li and
                  Yu Zhang and
                  Ming{-}Wei Chang},
  title        = {Conditional Adapters: Parameter-efficient Transfer Learning with Fast
                  Inference},
  booktitle    = {NeurIPS},
  year         = {2023}
}

@inproceedings{adapterdrop,
  author       = {Andreas R{\"{u}}ckl{\'{e}} and
                  Gregor Geigle and
                  Max Glockner and
                  Tilman Beck and
                  Jonas Pfeiffer and
                  Nils Reimers and
                  Iryna Gurevych},
  title        = {AdapterDrop: On the Efficiency of Adapters in Transformers},
  booktitle    = {{EMNLP} {(1)}},
  pages        = {7930--7946},
  publisher    = {Association for Computational Linguistics},
  year         = {2021}
}

@article{li2021prefix,
  title={Prefix-tuning: Optimizing continuous prompts for generation},
  author={Li, Xiang Lisa and Liang, Percy},
  journal={arXiv preprint arXiv:2101.00190},
  year={2021}
}

@article{lester2021power,
  title={The power of scale for parameter-efficient prompt tuning},
  author={Lester, Brian and Al-Rfou, Rami and Constant, Noah},
  journal={arXiv preprint arXiv:2104.08691},
  year={2021}
}

@article{wang2023multitask,
  title={Multitask prompt tuning enables parameter-efficient transfer learning},
  author={Wang, Zhen and Panda, Rameswar and Karlinsky, Leonid and Feris, Rogerio and Sun, Huan and Kim, Yoon},
  journal={arXiv preprint arXiv:2303.02861},
  year={2023}
}

@inproceedings{pissa,
  author       = {Fanxu Meng and
                  Zhaohui Wang and
                  Muhan Zhang},
  title        = {PiSSA: Principal Singular Values and Singular Vectors Adaptation of
                  Large Language Models},
  booktitle    = {NeurIPS},
  year         = {2024}
}

@inproceedings{dora,
  author       = {Shih{-}Yang Liu and
                  Chien{-}Yi Wang and
                  Hongxu Yin and
                  Pavlo Molchanov and
                  Yu{-}Chiang Frank Wang and
                  Kwang{-}Ting Cheng and
                  Min{-}Hung Chen},
  title        = {DoRA: Weight-Decomposed Low-Rank Adaptation},
  booktitle    = {{ICML}},
  publisher    = {OpenReview.net},
  year         = {2024}
}

@article{wu2024mixture,
  title={Mixture of lora experts},
  author={Wu, Xun and Huang, Shaohan and Wei, Furu},
  journal={arXiv preprint arXiv:2404.13628},
  year={2024}
}

@inproceedings{lora+,
  author       = {Soufiane Hayou and
                  Nikhil Ghosh and
                  Bin Yu},
  title        = {LoRA+: Efficient Low Rank Adaptation of Large Models},
  booktitle    = {{ICML}},
  publisher    = {OpenReview.net},
  year         = {2024}
}

@inproceedings{LiuD0PCCT24,
  author       = {Boan Liu and
                  Liang Ding and
                  Li Shen and
                  Keqin Peng and
                  Yu Cao and
                  Dazhao Cheng and
                  Dacheng Tao},
  title        = {Diversifying the Mixture-of-Experts Representation for Language Models with Orthogonal Optimizer},
  booktitle    = {{ECAI}},
  series       = {Frontiers in Artificial Intelligence and Applications},
  volume       = {392},
  pages        = {2966--2973},
  publisher    = {{IOS} Press},
  year         = {2024}
}

@inproceedings{hcsmoe,
  author       = {I{-}Chun Chen and
                  Hsu{-}Shen Liu and
                  Wei{-}Fang Sun and
                  Chen{-}Hao Chao and
                  Yen{-}Chang Hsu and
                  Chun{-}Yi Lee},
  title        = {Retraining-free Merging of Sparse MoE via Hierarchical Clustering},
  booktitle    = {{ICML}},
  publisher    = {OpenReview.net},
  year         = {2025}
}

@article{he2025efficiently_edit,
  title={Efficiently Editing Mixture-of-Experts Models with Compressed Experts},
  author={He, Yifei and Liu, Yang and Liang, Chen and Awadalla, Hany Hassan},
  journal={arXiv preprint arXiv:2503.00634},
  year={2025}
}

@misc{alpaca,
  author = {Rohan Taori and Ishaan Gulrajani and Tianyi Zhang and Yann Dubois and Xuechen Li and Carlos Guestrin and Percy Liang and Tatsunori B. Hashimoto },
  title = {Stanford Alpaca: An Instruction-following LLaMA model},
  year = {2023},
  publisher = {GitHub},
  journal = {GitHub repository},
  howpublished = {\url{https://github.com/tatsu-lab/stanford_alpaca}},
}

@article{arc,
  title={Think you have solved question answering? try arc, the ai2 reasoning challenge},
  author={Clark, Peter and Cowhey, Isaac and Etzioni, Oren and Khot, Tushar and Sabharwal, Ashish and Schoenick, Carissa and Tafjord, Oyvind},
  journal={arXiv preprint arXiv:1803.05457},
  year={2018}
}

@inproceedings{boolq,
  author       = {Christopher Clark and
                  Kenton Lee and
                  Ming{-}Wei Chang and
                  Tom Kwiatkowski and
                  Michael Collins and
                  Kristina Toutanova},
  title        = {BoolQ: Exploring the Surprising Difficulty of Natural Yes/No Questions},
  booktitle    = {{NAACL-HLT} {(1)}},
  pages        = {2924--2936},
  publisher    = {Association for Computational Linguistics},
  year         = {2019}
}

@inproceedings{piqa,
  author       = {Yonatan Bisk and
                  Rowan Zellers and
                  Ronan Le Bras and
                  Jianfeng Gao and
                  Yejin Choi},
  title        = {{PIQA:} Reasoning about Physical Commonsense in Natural Language},
  booktitle    = {{AAAI}},
  pages        = {7432--7439},
  publisher    = {{AAAI} Press},
  year         = {2020}
}

@article{winogrande,
  author       = {Keisuke Sakaguchi and
                  Ronan Le Bras and
                  Chandra Bhagavatula and
                  Yejin Choi},
  title        = {WinoGrande: an adversarial winograd schema challenge at scale},
  journal      = {Commun. {ACM}},
  volume       = {64},
  number       = {9},
  pages        = {99--106},
  year         = {2021}
}

@article{mobe,
  title={MoBE: Mixture-of-Basis-Experts for Compressing MoE-based LLMs},
  author={Chen, Xiaodong and Ha, Mingming and Lan, Zhenzhong and Zhang, Jing and Li, Jianguo},
  journal={arXiv preprint arXiv:2508.05257},
  year={2025}
}

@inproceedings{adamw,
  author       = {Ilya Loshchilov and
                  Frank Hutter},
  title        = {Decoupled Weight Decay Regularization},
  booktitle    = {{ICLR} (Poster)},
  publisher    = {OpenReview.net},
  year         = {2019}
}
\bibliographystyle{icml2026}

%%%%%%%%%%%%%%%%%%%%%%%%%%%%%%%%%%%%%%%%%%%%%%%%%%%%%%%%%%%%%%%%%%%%%%%%%%%%%%%
%%%%%%%%%%%%%%%%%%%%%%%%%%%%%%%%%%%%%%%%%%%%%%%%%%%%%%%%%%%%%%%%%%%%%%%%%%%%%%%
% APPENDIX
%%%%%%%%%%%%%%%%%%%%%%%%%%%%%%%%%%%%%%%%%%%%%%%%%%%%%%%%%%%%%%%%%%%%%%%%%%%%%%%
%%%%%%%%%%%%%%%%%%%%%%%%%%%%%%%%%%%%%%%%%%%%%%%%%%%%%%%%%%%%%%%%%%%%%%%%%%%%%%%
\newpage
\appendix
% \onecolumn

\section{Supplemental Experimental Results}

\subsection{Results on DeepSeek Model}

% deepseek_results_v4_complete
\begin{table*}[ht]
\small
\renewcommand\tabcolsep{5.5pt}
\renewcommand{\arraystretch}{1.0}
\centering
\setlength{\extrarowheight}{0pt}
\addtolength{\extrarowheight}{\aboverulesep}
\addtolength{\extrarowheight}{\belowrulesep}
\setlength{\aboverulesep}{0pt}
\setlength{\belowrulesep}{0pt}

\caption{Performance comparison on DeepSeek model across methods and tasks at different compression ratios.}
\label{tab:deepseek_results}
\begin{tabular}{c|c|c|cccccc}
\toprule
\hline \specialrule{0em}{1pt}{1pt}
    &                       &                       & \textbf{Math} & \textbf{Code} & \textbf{Common.} & \multicolumn{2}{c}{\textbf{Specialized Tasks}} &            \\ 
    \cmidrule(lr){4-4} \cmidrule(lr){5-5} \cmidrule(lr){6-6} \cmidrule(lr){7-8}
\multirow{-2}{*}{\textbf{Ratio}}  & 
  \multirow{-2}{*}{\textbf{Methods}} &
  \multirow{-2}{*}{\textbf{\# Params}} &
  GSM8K &
  HumanEval &
  8 Sub-Tasks &
  Intent &
  Translation &
  \multirow{-2}{*}{Average} 
  \\ \specialrule{0em}{1pt}{1pt} \hline \specialrule{0em}{1pt}{1pt}
0\% & Original             & -     & 3.0           & 53.9          & 61.8          & 0.0           & 12.2          & 26.2       \\
0\% & LoRA                 & 0.96B & 38.9          & 58.0          & 62.3          & 81.0          & 19.9          & 52.0      
\\ \specialrule{0em}{1pt}{1pt} \hline \specialrule{0em}{1pt}{1pt}
    & MoBE                 & 0.96B & \textbf{54.3} & 47.7          & \textbf{59.7} & \underline{80.2} & \textbf{27.0} & \textbf{53.8} \\
    & HC-SMoE              & 0.96B & 36.8          & 51.6          & 57.8          & 75.8          & \underline{26.9} & 49.8       \\
    & MC-SMoE              & 0.96B & \underline{48.4} & \underline{53.7} & 57.0       & 60.6          & 26.8          & 49.3       \\
\multirow{-4}{*}{30\%} &
  \cellcolor[HTML]{EFEFEF}\textbf{\MethodName{}} &
  \cellcolor[HTML]{EFEFEF}0.96B &
  \cellcolor[HTML]{EFEFEF}33.3 &
  \cellcolor[HTML]{EFEFEF}\textbf{58.1} &
  \cellcolor[HTML]{EFEFEF}\underline{59.2} &
  \cellcolor[HTML]{EFEFEF}\textbf{81.2} &
  \cellcolor[HTML]{EFEFEF}25.3 &
  \cellcolor[HTML]{EFEFEF}\underline{51.4}
  \\ \specialrule{0em}{1pt}{1pt} \hline \specialrule{0em}{1pt}{1pt}
    & MoBE                 & 0.96B & \textbf{50.9} & 40.6          & \textbf{57.6} & 71.2          & 24.5          & \underline{49.0} \\
    & HC-SMoE              & 0.96B & 43.0          & 47.8          & 55.1          & \underline{72.2} & 26.5       & 48.9       \\
    & MC-SMoE              & 0.96B & \underline{47.6} & \underline{50.2} & 54.7       & 61.8          & \underline{27.5} & 48.4       \\
\multirow{-4}{*}{40\%} &
  \cellcolor[HTML]{EFEFEF}\textbf{\MethodName{}} &
  \cellcolor[HTML]{EFEFEF}0.96B &
  \cellcolor[HTML]{EFEFEF}33.7 &
  \cellcolor[HTML]{EFEFEF}\textbf{55.8} &
  \cellcolor[HTML]{EFEFEF}\underline{57.3} &
  \cellcolor[HTML]{EFEFEF}\textbf{80.0} &
  \cellcolor[HTML]{EFEFEF}\textbf{27.7} &
  \cellcolor[HTML]{EFEFEF}\textbf{50.9}
  \\ \specialrule{0em}{1pt}{1pt} \hline \specialrule{0em}{1pt}{1pt}
    & MoBE                 & 0.96B & \underline{39.3} & 19.2          & 50.7          & 63.8          & 19.9          & 38.6       \\
    & HC-SMoE              & 0.96B & 34.3          & \underline{43.5} & \underline{53.0} & \underline{72.4} & \textbf{25.8} & \underline{45.8} \\
    & MC-SMoE              & 0.96B & \textbf{42.2} & 38.2          & 51.3          & 47.6          & 22.9          & 40.4       \\
\multirow{-4}{*}{50\%} &
  \cellcolor[HTML]{EFEFEF}\textbf{\MethodName{}} &
  \cellcolor[HTML]{EFEFEF}0.96B &
  \cellcolor[HTML]{EFEFEF}32.8 &
  \cellcolor[HTML]{EFEFEF}\textbf{51.5} &
  \cellcolor[HTML]{EFEFEF}\textbf{54.3} &
  \cellcolor[HTML]{EFEFEF}\textbf{79.4} &
  \cellcolor[HTML]{EFEFEF}\underline{24.0} &
  \cellcolor[HTML]{EFEFEF}\textbf{48.4}
  \\ \specialrule{0em}{1pt}{1pt} \hline
\bottomrule
\end{tabular}
\vspace{-0.2cm}
\end{table*}

% We further evaluate LightMoE on the DeepSeek-V2-Lite model with compression ratios of 30\%, 40\%, and 50\%. In addition to the first layer, the model originally contains 64 experts and one shared expert per layer.

To further verify the effectiveness of \MethodName{}, we extend our evaluation to the DeepSeek-V2-Lite model \cite{deepseekv2}, with 15.7B total and 2.4B active parameters. With the exception of the first layer, this model natively consists of 64 routed experts and 2 shared experts per layer. We maintain the same experimental setup and training configurations as those used for the OLMoE model to ensure consistency. The performance comparison across diverse tasks and compression ratios is presented in \cref{tab:deepseek_results}.

Overall, \MethodName{} demonstrates robust performance, achieving the best or second-best scores across most tasks. This advantage becomes particularly pronounced at higher compression ratios. While MoBE performs slightly better at a mild 30\% compression ratio, it suffers from a drastic performance collapse as compression increases, dropping to an average of 38.6\% at the 50\% ratio. In contrast, \MethodName{} maintains superior stability, securing the highest average performance at both 40\% and 50\% ratios. This highlights the robustness of our approach under aggressive compression constraints.

A critical observation in \cref{tab:deepseek_results} is the performance on the Math task. Notably, \MethodName{} (33.3\%) lags behind the standard LoRA baseline (38.9\%), while other compression methods like MoBE (54.3\%) and MC-SMoE (48.4\%) significantly outperform LoRA. We attribute this anomaly primarily to the rank capacity. Under the constraint of equal trainable parameters, \MethodName{} operates with a rank of 16, whereas LoRA uses a rank of 54, and other baselines utilize ranks exceeding 80. This trend suggests that complex mathematical reasoning is highly sensitive to low-rank bottlenecks. While our design ensures overall stability, the reduced rank limits the specific capacity for the math task in DeepSeek model. Future work could address this trade-off by exploring adaptive rank allocation strategies.

% \subsection{Examples for Tasks}

\subsection{Impact of Sample size on Importance Scoring}
% Analysis of Sample Size

\begin{table}[h]
    \centering
    \caption{Ablation study on the impact of sample size on OLMoE performance.}
    \label{tab:robustness_tokens}
    \begin{tabular}{c|cccc}
        \toprule
        \multirow{2}{*}{\textbf{Ratio}} & \multicolumn{4}{c}{\textbf{Number of Tokens}} \\
        & $2^{13}$ & $2^{15}$ & $2^{17}$ & $2^{19}$ \\
        \midrule
        30\% & 57.4 & 59.1 & \textbf{59.7} & 59.6 \\
        40\% & 55.1 & 56.6 & \underline{57.9} & \textbf{58.4} \\
        50\% & 53.9 & 54.5 & \underline{56.1} & \textbf{56.3} \\
        \bottomrule
    \end{tabular}
\end{table}

To evaluate the amount of data needed to select the less important experts for a task, we assess the model's performance on the Math task using varying token counts, ranging from  $2^{13}$ to $2^{19}$.
As shown in \cref{tab:robustness_tokens}, we observe that the performance improves as the number of tokens increases from $2^{13}$ to $2^{17}$, highlighting the necessity of adequate data for accurate expert selection. However, performance gains saturate as the sample size reaches $2^{17}$. Specifically, increasing the sample size from $2^{17}$ to $2^{19}$ yields minimal gains across all compression ratios. 
This indicates that the sample size of $2^{17}$ is sufficiently large to identify the less important experts.

\subsection{Sensitivity Analysis of Adaptive Thresholding Hyperparameters}

\begin{table}[ht]
    \centering
    \setlength{\tabcolsep}{5pt} % 紧凑的列间距
    \caption{Sensitivity analysis of the clipping range parameter $\text{max}\Delta$ for OLMoE on the Math task.}
    \label{tab:sensitivity_clipping}
    \begin{tabular}{c|ccccccc}
        \toprule
        \multirow{2}{*}{\textbf{Ratio}} & \multicolumn{7}{c}{\textbf{Clipping Range $\max\Delta$}} \\
        & 0 & 0.1 & 0.15 & 0.2 & 0.25 & 0.3 & $\infty$ \\
        \midrule
        30\% & 58.1 & 59.4 & \textbf{59.7} & \textbf{59.7} & 59.3 & 58.6 & 58.6 \\
        50\% & \textbf{57.1} & 55.6 & 55.8 & \underline{56.1} & 54.1 & 54.3 & 54.3 \\
        \bottomrule
    \end{tabular}
\end{table}

To provide a systematic justification for the hyperparameter choice in our adaptive thresholding mechanism, we introduce a parameter $\text{max}\Delta$ to control the bounds of the layer-specific thresholds. Specifically, the minimum and maximum thresholds are defined as:
\begin{equation}
    p_{\min} = (1-\text{max}\Delta)\hat{p}, \quad p_{\max} = (1+\text{max}\Delta)\hat{p}.
\end{equation}
We evaluate the model performance on the Math task with $\text{max}\Delta \in \{0, 0.1, 0.15, 0.2, 0.25, 0.3, \infty\}$. Notably, $\text{max}\Delta=0$ is equivalent to the "Uniform" strategy, while $\text{max}\Delta=\infty$ removes the clipping bounds entirely.

The results presented in \cref{tab:sensitivity_clipping} highlight two key observations. First, at a moderate 30\% compression ratio, our adaptive approach significantly outperforms the uniform baseline. Introducing a reasonable clipping range improves performance from 58.1\% to 59.7\%, confirming the benefit of varying thresholds based on layer importance. Second, the results demonstrate that moderate clipping is essential. Removing the bounds entirely leads to performance degradation compared to the optimal range (e.g., dropping from 59.7\% to 58.6\% at the 30\% ratio). This suggests that while flexibility is beneficial, preventing the thresholds from becoming too extreme is necessary for stability. 

In conclusion, values between 0.15 and 0.2 consistently yield robust performance across different compression ratios. This validates our default choice of $\text{max}\Delta=0.2$ as a systematic and effective setting that balances adaptivity with numerical stability.

\subsection{Impact of Group Size}

\begin{figure*}[ht]
    \centering
      \includegraphics[width=0.85\textwidth]{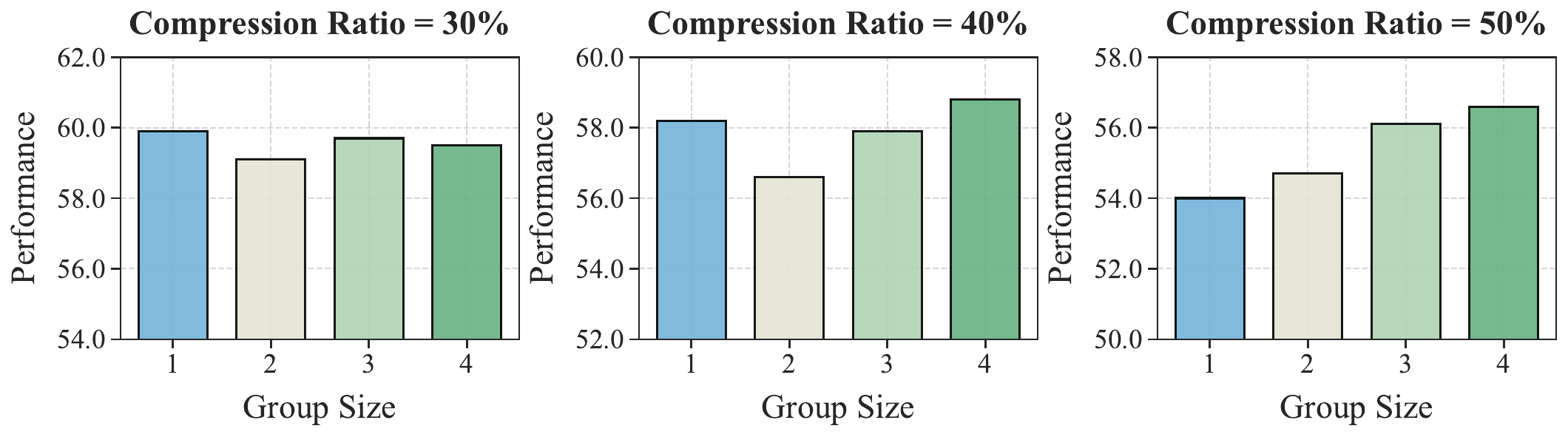}
    \caption{Comparison of different group sizes at different compression ratios for OLMoE on the Math task.}
    \label{fig:group_size}
    % \vspace{-0.4cm}
\end{figure*}

We explore the impact of incrementally increasing the group size of \MethodName{} from 1 to 4. \cref{fig:group_size} illustrates performance across various group sizes and compression ratios on the Math task.
At a mild 30\% compression ratio, the performance of all group sizes vary by less than 0.9\%, showing that multiple shared bases has little effect when ample parameters remain.

In contrast, as compression becomes more aggressive, the benefit of larger group sizes becomes pronounced. In particular, at a 50\% compression ratio, performance improves with increasing group size. This trend suggests that under constrained parameter budgets, constructing multiple shared bases by grouping is crucial for preserving the model’s knowledge and capabilities.

% \subsection{Expert Grouping Strategies}

% \begin{table}[h]
%     \centering
%     \caption{Comparison of expert grouping strategies for OLMoE at different compression ratios.}
%     \label{tab:grouping_comparison}
%     \begin{tabular}{c|cc}
%         \toprule
%         \textbf{Ratio} & \textbf{K-means} & \textbf{Dominant (Ours)} \\
%         \midrule
%         30\% & \textbf{60.0} & 59.7 \\
%         40\% & 56.1 & \textbf{57.9} \\
%         50\% & 54.9 & \textbf{56.1} \\
%         \bottomrule
%     \end{tabular}
% \end{table}

% To explore different ways of grouping experts, we compare our method with the K-means clustering strategy, which is adapted from HC-SMoE \cite{hcsmoe}. 
% As shown in \cref{tab:grouping_comparison}, both strategies perform similarly at a mild 30\% compression ratio. However, as the compression becomes more aggressive, our method demonstrates a clear advantage. At the 40\% and 50\% ratios, our approach significantly outperforms K-means, achieving scores of 57.9 and 56.1 compared to 56.1 and 54.9, respectively.

% The likely reason for this difference lies in how the groups are formed. Our method focuses on preserving the "dominant" experts and grouping similar experts around them. K-means, on the other hand, computes average centroids for all experts. When the compression is high, K-means might inadvertently merge important experts into a general group, causing the model to lose specific knowledge. This suggests that explicitly preserving dominant experts is key to maintaining performance under high compression.

\subsection{Exploration of Annealing Schedules}

\begin{figure}[h]
    \centering
    \includegraphics[width=1.0\linewidth]{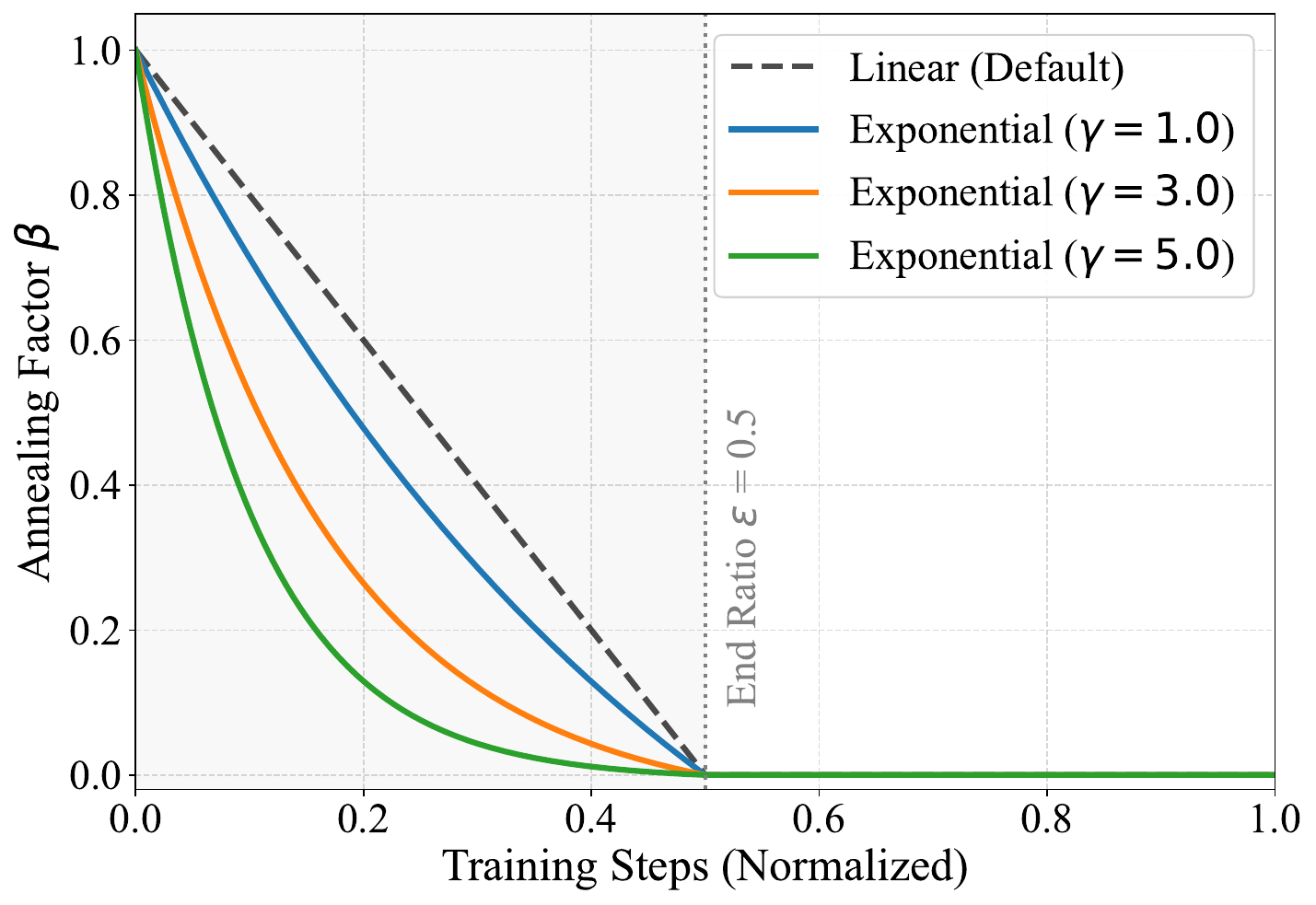} % 假设你保存的文件名是这个
    \caption{Comparison of different annealing schedules. The dashed line represents the linear decay strategy, while solid lines depict exponential decay strategies with varying $\gamma$. All schedules converge to zero at the end ratio $\epsilon$.}
    \label{fig:annealing_curves}
\end{figure}

To investigate the impact of the annealing schedule on model performance, we compare our default linear decay strategy against an exponential decay strategy. While the linear strategy decreases $\beta$ at a constant rate, the exponential strategy introduces a curvature controlled by a hyperparameter $\gamma$. Formally, maintaining consistency with the notation in \cref{sec:replacing}, the annealing factor $\beta$ under the exponential schedule is defined as:
\begin{equation}
    \beta = \max \left( \frac{e^{-\gamma \tau} - e^{-\gamma}}{1 - e^{-\gamma}}, 0 \right), \quad \text{where } \tau = \frac{t}{\epsilon T}.
\end{equation}
Here, $\tau$ represents the normalized training progress relative to the annealing period, and $\gamma$ modulates the decay shape. A larger $\gamma$ causes $\beta$ to drop more rapidly initially, whereas a smaller $\gamma$ maintains a higher value for a longer duration. Visualizations of these decay trajectories are provided in \cref{fig:annealing_curves}.

\begin{table}[h]
    \centering
    \caption{Comparison of linear and exponential annealing schedules for OLMoE on the Math task.}
    \label{tab:annealing_schedule}
    \begin{tabular}{c|cccc}
        \toprule
        \multirow{2}{*}{\textbf{Ratio}} & \multirow{2}{*}{\textbf{Linear}} & \multicolumn{3}{c}{\textbf{Exponential Decay}} \\
        & & $\gamma=1.0$ & $\gamma=3.0$ & $\gamma=5.0$ \\
        \midrule
        30\% & \underline{59.7} & \textbf{59.8} & 59.6 & 58.7 \\
        40\% & 57.9 & 56.1 & \underline{58.2} & \textbf{58.4} \\
        50\% & \textbf{56.1} & 54.5 & 54.4 & \underline{56.0} \\
        \bottomrule
    \end{tabular}
\end{table}

We evaluate this strategy with $\gamma$ values of 1.0, 3.0, and 5.0. The results on the Math task are summarized in \cref{tab:annealing_schedule}. Overall, the performance differences between the linear and exponential schedules are marginal. However, the linear schedule demonstrates remarkable robustness across different compression levels. It achieves the best performance at the aggressive 50\% compression ratio and ranks second-best at the 30\% ratio. Even at 40\%, where it falls slightly behind the exponential setting, the gap is negligible.

In contrast, the exponential strategy proves sensitive to the hyperparameter $\gamma$. A lower $\gamma$ performs well at low compression but degrades at high compression, whereas a higher $\gamma$ shows the opposite trend. No single $\gamma$ value yields optimal results across all scenarios. Consequently, we adopt the Linear schedule as a simple yet sufficiently effective solution, as it delivers consistently high performance without the need for additional hyperparameter tuning.

\section{Efficiency Analysis}

\begin{table}[h]
    \centering
    \caption{Inference efficiency comparison on the Math task under different compression ratios.}
    \label{tab:inference_efficiency}
    % 使用较小字体以适应双栏宽度，同时调整列间距
    % \small
    \setlength{\tabcolsep}{3.5pt} 
    \begin{tabular}{c|cc|cc}
        \toprule
        \multirow{2}{*}{\textbf{Ratio}} & \textbf{Total} & \textbf{Memory} & \textbf{MoE Active} & \textbf{Latency} \\
         & \textbf{Params} & \textbf{Usage} & \textbf{Params} & (ms/token) \\
        \midrule
        0\%  & 6.92 B & 12.89 GB & 0.81 B & 8.09 \\
        30\% & 4.82 B & 9.09 GB  & 0.77 B & 8.71 \\
        40\% & 4.14 B & 7.85 GB  & 0.73 B & 8.32 \\
        50\% & 3.47 B & 6.63 GB  & 0.70 B & 8.01 \\
        \bottomrule
    \end{tabular}
\end{table}

To assess the inference efficiency of our approach, we evaluate the model across four key metrics: total parameter count, GPU memory usage, MoE average active parameters per token, and inference latency. The results for the OLMoE model on the Math task are summarized in \cref{tab:inference_efficiency}.

As the compression ratio increases, we observe a substantial reduction in memory requirements. Specifically, the memory footprint decreases by nearly half, dropping from 12.89 GB in the original model to 6.63 GB at the 50\% compression ratio. In terms of computational cost, the decrease in the number of parameters activated within the MoE layers is less pronounced. This is expected, as our method primarily targets and compresses redundant experts that are less frequently activated, while preserving the critical experts that contribute most to the active parameter count. Moreover, although replacing experts with shared bases and adapters introduces slight architectural complexity, our method maintains an inference latency comparable to the original model. These findings demonstrate that \MethodName{} effectively reduces parameter redundancy and memory usage without compromising inference efficiency.

% \section{Impact of Annealing on Training Loss Trajectories}
\section{Training Dynamics Analysis}
\label{sec:training loss}

\begin{figure}[ht]
\centering
\includegraphics[width=1.0\linewidth]{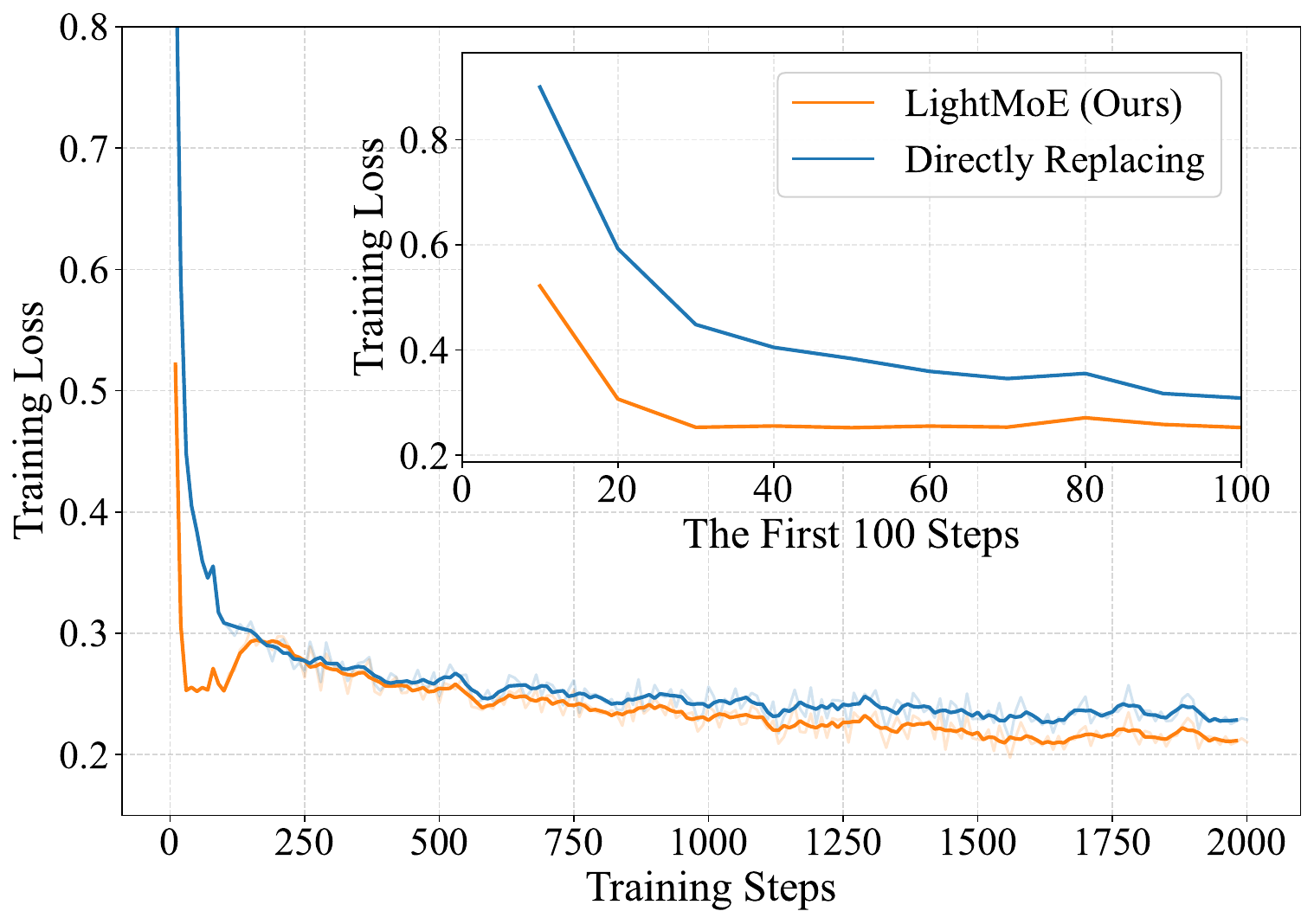}
\caption{Training loss trajectories of LightMoE versus the Directly Replacing baseline on the Math task with OLMoE at a 50\% compression ratio.}
\label{fig:training_loss}
\end{figure}

To validate the intuition behind our annealing strategy, we visualize the training dynamics of LightMoE compared to the Directly Replacing baseline in \cref{fig:training_loss}.

As illustrated in the inset of \cref{fig:training_loss}, the Directly Replacing baseline suffers from a severe optimization instability, evidenced by a sharp initial spike in loss. This indicates that the abrupt parameter substitution destabilizes the model, forcing the optimization trajectory to recover from a high-loss, sub-optimal state. In contrast, LightMoE ensures a smooth start with minimal loss, effectively mitigating the destructive initial shock.

As training progresses and $\beta$ decays, LightMoE exhibits a slight rise in loss. This phenomenon represents the critical transfer of capabilities from the original experts to the compressed modules. Effectively, the annealing phase acts as a warm-up period, allowing the compressed modules to implicitly align with the original experts' behavior.

Crucially, despite this transient rise, the loss trajectory of LightMoE remains lower than that of the baseline throughout the training process. This suggests that the smooth transition keeps the model in a better region of the parameter space, avoiding the suboptimal local minima that the baseline falls into.

Moreover, although our current strategy is effective, the temporary rise in loss implies a brief gap in capability during the transition. Future work could explore more adaptive schedules and better initialization methods to further mitigate this fluctuation, thereby enhancing the training stability of the expert replacing paradigm.
% Moreover, although our current strategy is effective, the temporary rise in loss suggests a brief capability gap during the transition. Future work could explore more adaptive schedules to further reduce this fluctuation and improve training stability.

\section{Base Threshold Settings}

\begin{table}[h]
    \centering
    % \small % 使用较小字体以适应单栏
    % \footnotesize 
    \setlength{\tabcolsep}{4pt} % 略微缩减列间距以防溢出
    \caption{Base threshold $\hat{p}$ settings calibrated for OLMoE and DeepSeek models. The average number of selected experts per layer (Avg \# Exp.) corresponds to the target compression ratio.}
    \label{tab:base_thresholds}
    \begin{tabular}{c c | c c c c c}
        \toprule
        \textbf{Ratio} & \textbf{Avg \#} & \multicolumn{5}{c}{\textbf{Base Threshold $\hat{p}$}} \\
        \textbf{(\%)} & \textbf{Exp.} & Math & Code & Common. & Intent & Trans. \\
        \midrule
        \multicolumn{7}{c}{\textit{\textbf{Model: OLMoE-1B-7B-SFT }}} \\
        \midrule
        30 & $\approx$ 24 & 0.110 & 0.095 & 0.185 & 0.055 & 0.036 \\
        40 & $\approx$ 31 & 0.180 & 0.159 & 0.275 & 0.098 & 0.059 \\
        50 & $\approx$ 38 & 0.268 & 0.240 & 0.382 & 0.168 & 0.092 \\
        \midrule
        \multicolumn{7}{c}{\textit{\textbf{Model: DeepSeek-V2-Lite}}} \\
        \midrule
        30 & $\approx$ 26 & 0.160 & 0.200 & 0.192 & 0.085 & 0.058 \\
        40 & $\approx$ 33 & 0.240 & 0.285 & 0.280 & 0.142 & 0.098 \\
        50 & $\approx$ 40 & 0.330 & 0.385 & 0.380 & 0.215 & 0.146 \\
        \bottomrule
    \end{tabular}
\end{table}

To facilitate reproducibility, we provide the specific hyperparameter configurations used to achieve the target compression ratios for both the OLMoE and DeepSeek models. In the LightMoE framework, the global compression ratio is controlled by the base threshold $\hat{p}$ defined in \cref{equ:adaptive threshold}. This threshold determines the subset of experts selected for replacement. A higher $\hat{p}$ results in more experts being replaced, thereby increasing the compression ratio.

Since the distribution of expert importance scores varies across domains, the base threshold required to select these experts differs. Therefore, for each task and target compression ratio, we determine the optimal $\hat{p}$ via a binary search process on the calibration dataset.
\cref{tab:base_thresholds} presents these $\hat{p}$ values for the five evaluated tasks, alongside the corresponding average number of experts selected per layer.

A distinct pattern emerges from the results. For preservation tasks, including Math, Code, and Commonsense Reasoning, the base thresholds are relatively high. This suggests that in these domains, even the less critical experts selected for replacement retain a moderate level of activation and contribution. Conversely, for adaptation tasks like Intent Recognition and Translation, the thresholds are significantly lower. This indicates that the expert activation in these specialized downstream tasks is highly sparse, allowing a large number of redundant experts to be identified even with a very low threshold.

% \section{You \emph{can} have an appendix here.}

% You can have as much text here as you want. The main body must be at most $8$
% pages long. For the final version, one more page can be added. If you want, you
% can use an appendix like this one.

% The $\mathtt{\backslash onecolumn}$ command above can be kept in place if you
% prefer a one-column appendix, or can be removed if you prefer a two-column
% appendix.  Apart from this possible change, the style (font size, spacing,
% margins, page numbering, etc.) should be kept the same as the main body.
%%%%%%%%%%%%%%%%%%%%%%%%%%%%%%%%%%%%%%%%%%%%%%%%%%%%%%%%%%%%%%%%%%%%%%%%%%%%%%%
%%%%%%%%%%%%%%%%%%%%%%%%%%%%%%%%%%%%%%%%%%%%%%%%%%%%%%%%%%%%%%%%%%%%%%%%%%%%%%%

\end{document}